\documentclass[10pt,twocolumn,letterpaper]{article}

\usepackage[pagenumbers]{cvpr} 

\definecolor{cvprblue}{rgb}{0.21,0.49,0.74}
\usepackage[pagebackref,breaklinks,colorlinks,allcolors=cvprblue]{hyperref}
\usepackage[hyphens]{url}
\usepackage{algorithm}
\usepackage{algorithmic}
\usepackage{amsmath}
\usepackage{cancel}
\newtheorem{problem}{Problem}

\newtheorem{definition}{Definition}

\newlength\savedwidth
\usepackage{booktabs}
\usepackage{bm}
\newcommand{\piindex}{$\pi^*$-index }
\usepackage{physics}
\usepackage{amssymb}
\usepackage{amsmath}
\usepackage{cancel}
\usepackage{mathtools}
\mathtoolsset{showonlyrefs}

\title{Explaining Object Detectors via Collective Contribution of Pixels}

\author{
Toshinori Yamauchi$^{1}$ \and
Hiroshi Kera$^{1,2}$ \and
Kazuhiko Kawamoto$^{1}$ \and
$^{1}$Chiba University \quad $^{2}$National Institute of Informatics\\
{\tt\small t.yamauchi@chiba-u.jp, kera@chiba-u.jp, kawa@faculty.chiba-u.jp}
}

\begin{document}
\maketitle

\begin{abstract}
Visual explanations for object detectors are crucial for enhancing their reliability.
Object detectors identify and localize instances by assessing multiple visual features collectively. 
When generating explanations, 
overlooking these collective influences in detections may lead to missing compositional cues or capturing spurious correlations.
However, 
existing methods typically focus solely on individual pixel contributions, 
neglecting the collective contribution of multiple pixels.
To address this limitation, 
we propose a game-theoretic method based on Shapley values and interactions to
explicitly capture both individual and collective pixel contributions.
Our method provides explanations for  
both bounding box localization and class determination, 
highlighting regions crucial for detection.
Extensive experiments demonstrate that the proposed method identifies important regions more accurately than state-of-the-art methods. 
The code is available at \url{https://github.com/tttt-0814/VX-CODE}.
\end{abstract}

\section{Introduction}
\label{sec:intro}
Object detectors are applied in safety-critical applications such as autonomous driving~\cite{intro1,intro2} and medical imaging~\cite{intro3,intro4}.
To make object detectors more reliable, 
it is crucial to understand which pixels in an image 
influence the detection. 
Such analysis also helps identify potential biases in the training data and improve their robustness.
Several methods have been proposed to visualize image pixels that strongly contribute to detection~\cite{drise,odam,l-crp,bsed, ssgcam++}. 

\par
\begin{figure}[t]
    \centering
    \includegraphics[width=\linewidth]{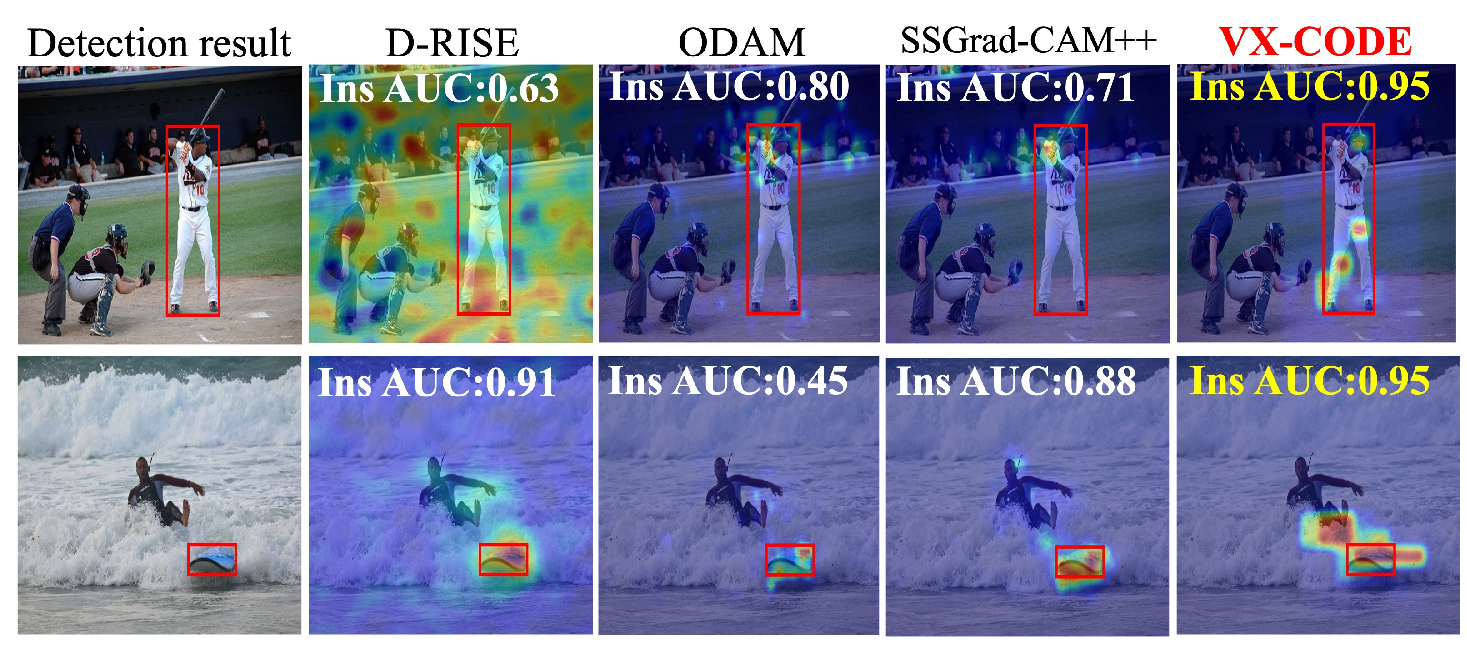}
    \caption{
    Generated heat maps by D-RISE, ODAM, SSGrad-CAM++, and VX-CODE.
    Each heat map shows the insertion AUC (Ins AUC), which
    measures faithfulness (higher is better).
    While previous methods highlight dominant features (e.g., hand or surfboard),
    VX-CODE captures collective contributions across multiple features (e.g., leg+head or surfboard+sea) through interactions. 
    }
    \label{fig:intro}
\end{figure}

However, 
existing methods overlook the \textit{collective contributions} of image pixels, 
which may lead to missing compositional cues or capturing spurious correlations.
As shown in Fig.~\ref{fig:intro}, state-of-the-art methods, such as ODAM~\cite{odam} and SSGrad-CAM++~\cite{ssgcam++}, 
often fail to capture these collective effects: these methods mainly highlight the hands (top) or the surfboard (bottom), while overlooking pixels that indirectly affect the bounding-box size or the predicted class. 
This limitation becomes critical when explaining biased models or failure cases, which often arise from spurious correlations (cf.~Figs.~\ref{fig:heatmap_bias_auc} and~\ref{fig:failer_case}). 
These issues stem from evaluating pixel contributions independently, which ignores subtle but important cues for the detector's decision.

In this study, 
we propose \textbf{V}isual e\textbf{X}planation of \textbf{CO}llective contributions for object \textbf{DE}tectors ({VX-CODE}), an explanation method for object detectors that explicitly accounts for the collective contributions of image pixels.
VX-CODE achieves more accurate identification of important pixels that influence object detection (i.e., the bounding box localization and class prediction); 
see Fig.~\ref{fig:intro}. 
VX-CODE leverages game-theoretic metrics:
the Shapley value~\cite{shapley} to measure individual contributions
and interactions~\cite{interaction} to measure  collective contributions of multiple pixels.
Although the direct computation of these metrics scales exponentially with the number of pixels (or patches), 
we adopt a greedy patch selection strategy that 
efficiently incorporates these metrics into the explanation process.
Specifically, image patches are greedily inserted into an empty image (or deleted from the original image) to maximize (or minimize) a detection score that reflects both bounding-box and class predictions.

\par
Our extensive experiments with widely used object detectors~\cite{detr,fast_r_cnn} and benchmark datasets~\cite{coco, voc} demonstrate that VX-CODE provides more faithful explanations, both qualitatively and quantitatively, than existing state-of-the-art methods.
In particular, the insertion and deletion AUC scores 
improve by up to approximately 19\,\%.
The experiments also cover bias models (Sec.~\ref{sec:problem_setup} and App.~\ref{sec:more_bias_model}), 
failure cases (Sec.~\ref{sec:problem_setup} and App.\ref{sec:failer_cases}), 
the pointing game (App.~\ref{sec:pg}), 
and explanations for bounding-box generation (App.~\ref{sec:box_explanation}).
In all such settings, 
VX-CODE outperforms state-of-the-art methods, 
highlighting the importance of accounting for the collective contributions of pixels. 

Our contributions are summarized as follows.
\begin{itemize}
  \item {We propose VX-CODE, the first efficient method that leverages \textit{collective contributions} of pixels to identify essential pixels that determine the output of an object detector (i.e., bounding box and class prediction).}
  \item {We develop and analyze the efficient integration of pixel synergies using game-theoretic indices, particularly Shapley values and interactions, along with their efficient variants. 
  Further, we introduce the $\pi^*$-index, a new index that provides a theoretical interpretation of the greedy framework from a sequential coalitional game perspective.
  }
  \item {We validate VX-CODE through extensive experiments.
  Existing methods fail to capture information that is critical yet not dominant, as they assess only individual contributions. In contrast, VX-CODE accounts for the collective influence of such features on detection, leading to more faithful and informative explanations.
}
\end{itemize}

\section{Related Work}
\label{sec:related_work}
Most visual explanation methods have been developed for classification models, 
while object detectors have been less studied. 
Existing methods for object detectors are often adaptations of techniques originally designed for classification.
We summarize representative methods for classification in App.~\ref{sec:related_classification},
and focus on those specifically developed for object detectors~\cite{faster_r_cnn, ssd, yolo, fcos,detr}.
These methods can be broadly categorized into three groups: gradient-based methods, class activation map (CAM)-based methods, and perturbation-based methods.

\par
Gradient-based methods compute instance-specific importance by propagating gradients through the network~\cite{odsmoothgrad, l-crp, odam}.
ODAM~\cite{odam} multiplies feature and gradient maps to estimate importance.
CAM-based methods, 
like SSGrad-CAM~\cite{ssgcam} and SSGrad-CAM++~\cite{ssgcam++},
extend Grad-CAM~\cite{gradcam} by incorporating spatial information.
Perturbation-based methods measure changes in the detector’s output when parts of the input images are masked~\cite{drise, d-close, vps}.
D-RISE~\cite{drise}, an extension of RISE~\cite{RISE},
masks inputs randomly, while
BSED~\cite{bsed} and FSOD~\cite{fsod} use Shapley values to estimate per-pixel or per-region importance.

\par
These methods only capture the individual contributions of each pixel and fail to model interactions among pixels.
Recent work in classification, such as PredDiff~\cite{pred-diff} and MoXI~\cite{moxi}, shows that modeling collective contributions can yield more faithful explanations. 
As discussed in Sec.~\ref{sec:intro}, 
pixel interactions are also crucial for understanding object detectors.
To this end, 
we propose VX-CODE, a perturbation-based method that captures collective contributions through a greedy patch selection strategy.
Unlike previous patch selection methods~\cite{SAG,REX,vps}, 
VX-CODE is the first to analyze this strategy through the lens of Shapley values and interactions.
This perspective provides new insights into explaining object detectors from a sequential coalitional game perspective.

\section{Preliminaries}
\label{sec:preliminaries}
We use Shapley values~\cite{shapley} to measure individual contributions and interactions~\cite{interaction} to capture collective effects in coalitional games.
Let $N=\{1,...,n\}$ be player indices, $\mathcal{P}(N){=}\{S \mid S\subseteq N\}$ the power set, and 
$f:\mathcal{P}(N) \rightarrow \mathbb{R}$ a reward. 

The Shapley value averages the marginal gain of a player over all contexts.
\begin{definition}\label{def:shapley}
    The \textbf{Shapley value} $\phi (i|N)$ of player $i \in N$ is defined by
    \begin{align}
    \phi(i|N) \stackrel{\mathrm{def}}{=}\hspace{-0.7em}  \sum_{S \subseteq N \setminus \{i\}}\hspace{-0.7em} P(S | N \setminus \{i\})[f(S \cup \{i\}) - f(S)],
    \label{eq:shapley}
    \end{align}
where $P(A | B) = \frac{(|B| - |A|)!|A|!}{(|B| + 1)!}$ for sets $A$ and $B$, and $|\cdot|$ is the size of a set.
\end{definition}
The Shapley value $\phi (i|N)$ is the average marginal reward when player $i$ joins $S\subseteq N\setminus\{i\}$. 

\par
The interaction of two players is defined by a difference between the Shapley values when both of them exist and when one of them is absent.
\begin{definition}\label{def:interaction}
    The \textbf{interaction} of two players $i, j \in N$ is defined by
    \begin{align}
I(i,j|N) \stackrel{\mathrm{def}}{=} \phi(S_{ij} | N') - \phi(i | N \setminus \{j\}) - \phi(j | N \setminus \{i\}),
\label{eq:interaction}
\end{align}
where players $i$ and $j$ are regarded as a single entity $S_{ij}=\{i,j\}$, 
and $N' = N \setminus \{i,j\} \cup \{S_{ij}\}$.
\end{definition}
Importantly, 
when two patches $i$ and $j$ encode nearly the same information and the merged player adds no more than either one alone, i.e.,
$\phi(S_{ij}\mid N') \approx \phi(i\mid N\setminus\{j\}) \approx \phi(j\mid N\setminus\{i\}) =: m > 0$,
the interaction becomes negative.
This characteristic is crucial for identifying a minimal,
non-redundant set of patches that well explain object detectors.
If one selects only patches with high Shapley values as in prior study, 
patches containing assistive information (e.g., background) may be overlooked, as they typically contribute less individually compared to those containing primary information (e.g., the body of objects).

\par
An object detector $\mathcal{D}: x \mapsto \{(L_i, P_i)\}_{i=1}^m$ maps an image $x$ to a large set of detection proposals $\{(L_i, P_i)\}_{i=1}^m$. 
Each proposal $(L_i, P_i)$ includes a bounding box 
$L_i = (x_1^i, y_1^i, x_2^i, y_2^i)$ 
and a $C$-class probability vector $P_i \in [0, 1]^C$. 
Post-processing then selects a subset of proposals as final detections. 
In this setup, image patches $N$ of an image $x$ serve as players.
Player absence is implemented by applying corruptions such as local zero masking~\citep{drise} or blurring~\citep{blur1}. 
For a subset of patches $S \subseteq N$, 
we denote by $x_S$ the associated partially corrupted image.
Let $(L^x, P^x)$ be one of the final detections for the full image $x$. 
We adopt the following reward function from~\cite{drise}.
\begin{align}
    f(\mathcal{D}(x_S); (L^x, P^x)) = \underset{(L, P) \in \mathcal{D}(x_S)} {\operatorname{max}} \mathrm{IoU}(L^x,L) \cdot \frac{P^x \cdot P}{\|P^x\| \|P\|},
    \label{eq:reward}
\end{align}
where $\mathrm{IoU}(L^x,L)$ measures localization and the cosine term measures classification agreement.
Note that in general, $(L^x, P^x) \notin \mathcal{D}(x_S)$ unless $S=N$.
We provide a conceptual illustration of this computation in App.~\ref{sec:reward_concept},
validate it in App.~\ref{sec:limitation}, 
and further analyze the effect of the reward formulation on the resulting explanations in App.~\ref{sec:change_reward}.

\begin{figure}[t]
    \centering
    \includegraphics[width=\linewidth]{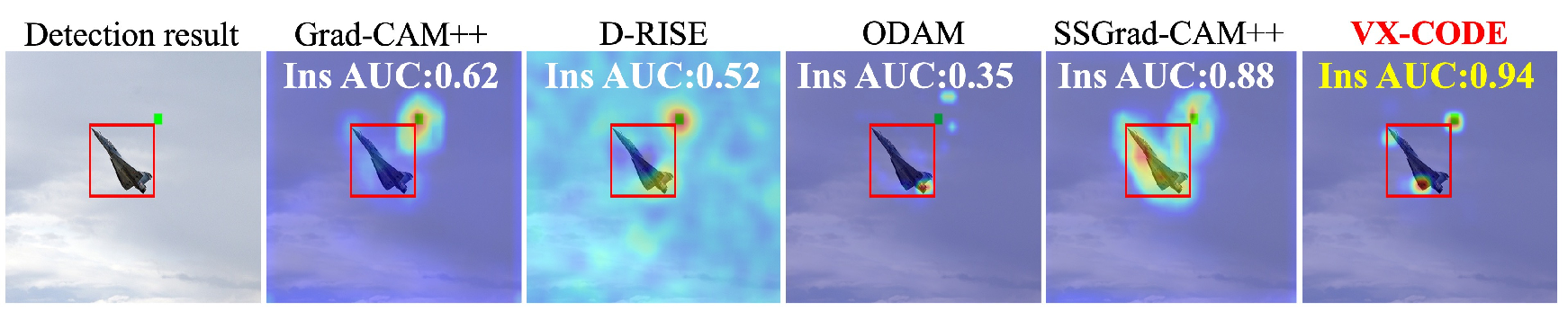}
    \caption{Comparison of visualizations generated by existing methods and VX-CODE with patch insertion ($r=1$) for an aeroplane detected using the biased model.}
    \label{fig:heatmap_bias_auc}
\end{figure}

\section{Approach}
\label{sec:approach}

\label{sec:problem_setup}
We identify image patches that contribute significantly to object detection.
Formally, we define the problem as follows. 
\begin{problem}\label{problem:problem_insertion}
Let $N$ be the index set of all patches in an image $x$, and
let $f:\mathcal{P}(N) \to [0, 1]$ be the reward function defined in 
Eq.~\eqref{eq:reward}. 
Find a subset $S_k \subset N$ of predesignated size $k$ such that 
    \begin{align}
        \text{(Insertion setup)} \quad S_k &= \underset{S\subseteq N, |S|=k}{\mathrm{arg\,max}} \; f(S), \\
        \text{(Deletion setup)}\quad S_k &= \underset{S\subseteq N, |S|=k}{\mathrm{arg\,min}} \; f(N\setminus S). 
    \end{align}
\end{problem}

The insertion setup identifies the minimal information required for successful detection (e.g., part of the object body and its background), whereas the deletion setup finds more comprehensive information (e.g., the entire object body) whose removal leads to detection failure.

\par
Prior studies have addressed this problem using measures based on input gradients, class activation maps, or Shapley values.
However, these measures capture only the independent contribution of each patch. 
A collection of highly contributing patches does not necessarily form the most informative region.
For example, we illustrate two motivating examples to clarify why modeling collective contributions is essential for explaining object detectors.

\vspace{0.7em}
\noindent \textbf{Explaining biased model.}
We train a biased model on a dataset where 70\% of the images contain a class-agnostic square marker placed outside the upper-right corner of the bounding box. 
As a result, the trained model relies on both objects and the marker for detection.
In Fig.~\ref{fig:heatmap_bias_auc},
VX-CODE, 
which considers collective contributions, 
clearly identifies both cues.
Previous methods highlight only the square marker (Grad-CAM++ and D-RISE) or almost the entire bounding box (SSGrad-CAM++).
The AUC score quantitatively confirm these observations. 
See App.~\ref{sec:more_bias_model} for additional results.

\begin{figure}[t]
    \centering
    \includegraphics[scale=0.32]{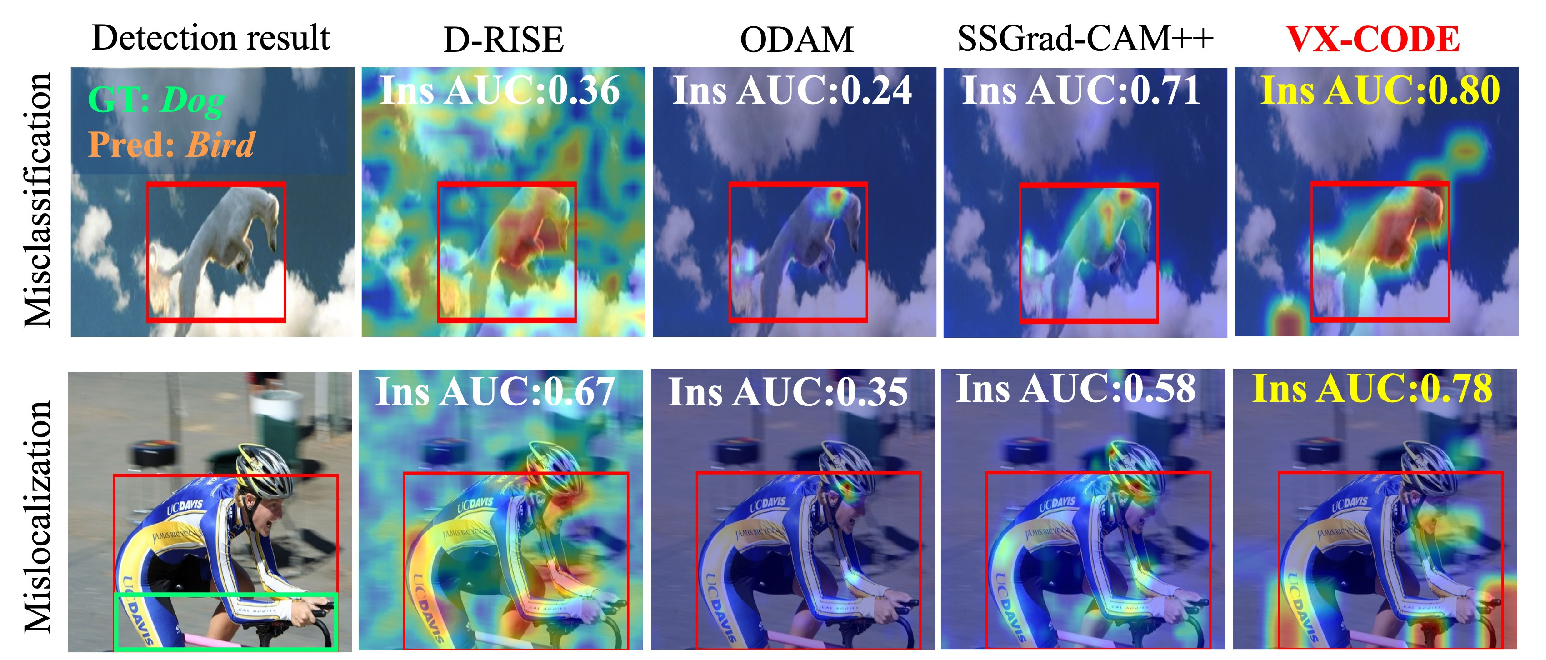}
    \caption{Comparison of visualizations generated by existing methods and VX-CODE with patch insertion ($r=1$) for failure cases. 
    In mislocalization, the green and red boxes indicate the ground truth and the prediction, respectively.}    
    \label{fig:failer_case}
\end{figure}

\vspace{0.7em}
\noindent \textbf{Explaining failure cases.}
Figure~\ref{fig:failer_case} presents two failure cases of a standard detection model. The first case is a misclassification of a dog as a bird due to the presence of sky, and the second is a mislocalization of the bicycle caused by the human. 
VX-CODE, considering collective contributions, correctly identifies the causes of these misdetections: both the sky and the human's body (e.g., face and legs) in the second.
ODAM and SSGrad-CAM++ highlight only the body or human-related features,
whereas D-RISE produces noisy explanations.
In both cases, the AUC score quantitatively supports these findings. 
See Sec.~\ref{sec:failer_cases} and App.~\ref{sec:eval_failer_case} for additional results.

\begin{figure}[t]
    \centering
    \includegraphics[scale=0.36]{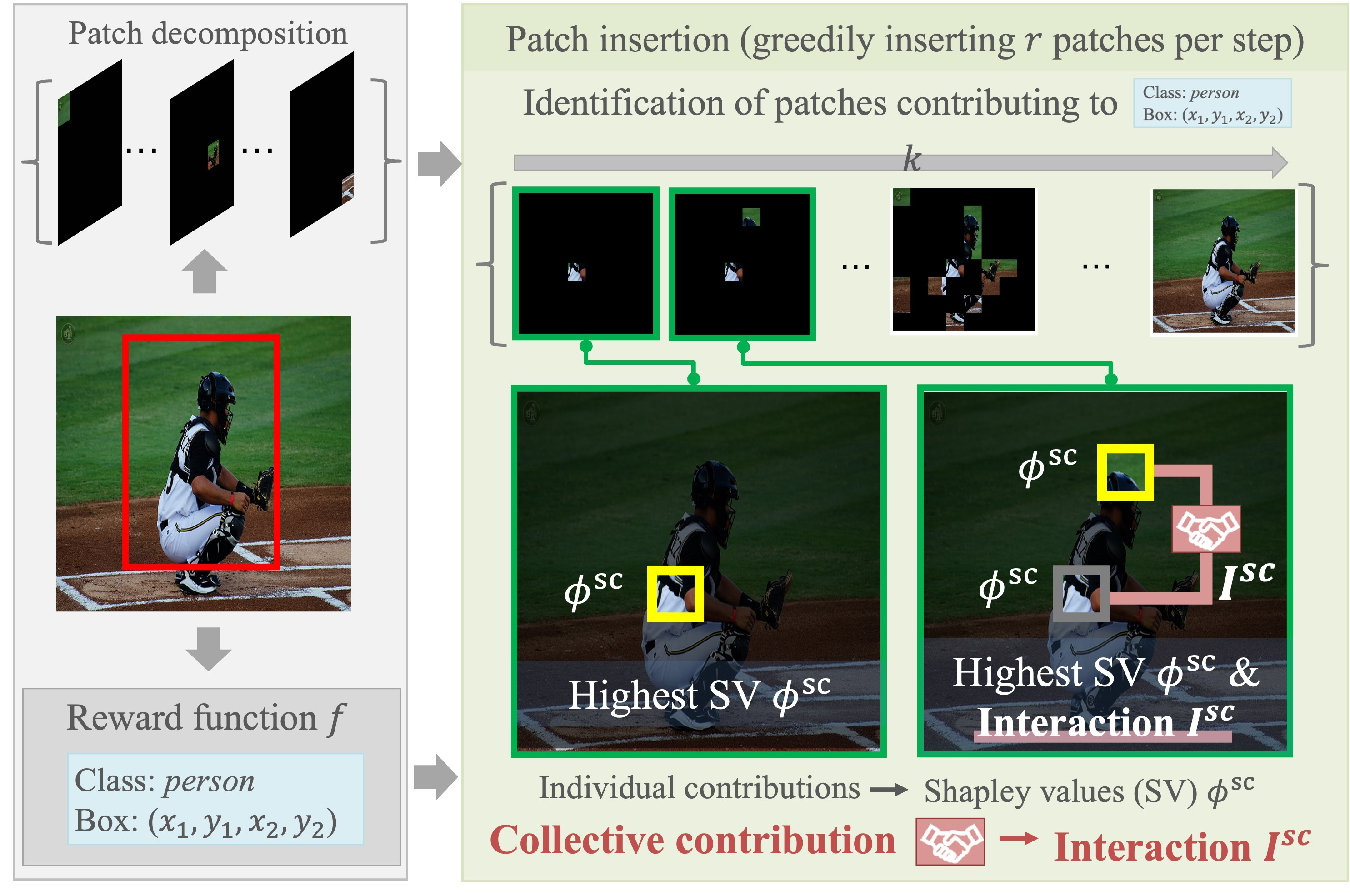}
    \caption{Overview of VX-CODE. The input image is divided into patches, 
    and $r$ patches are selected at each step. 
    This selection process considers not only individual contributions but also collective contributions through interaction.
    This figure illustrates the case of $r=1$, with step $k=1$ and $k=2$ in patch insertion.
    Gray-bordered patches represent the previously selected patches $\mathcal{B}_{k-1}$, 
    while yellow-bordered patches represent newly selected patches $B_{k}$.
    See App.~\ref{sec:pseudo_code} for detailed algorithms.}
    \label{fig:proposed_method}
\end{figure}

\subsection{Proposed method}
\label{sec:patch_insertion}
One of the main challenges in applying Shapley values and interactions is their high computational cost. 
To address this, we propose \emph{self-context} variants of Shapley values and interactions~\cite{moxi}, analyzed through our newly introduced $\pi^*$-index. 
The goal is not to approximate the original Shapley values and interactions, but to efficiently incorporate collective contributions into the explanation of object detectors (see App.~\ref{sec:limitation} for details).

\par
To this end,
we propose VX-CODE, an efficient heuristic to address Problem~\ref{problem:problem_insertion}. 
Figure~\ref{fig:proposed_method} provides an overview of the computation pipeline. 
Here, we focus on the insertion setup (patch insertion); 
the deletion setup (patch deletion) is methodologically similar and deferred to App.~\ref{sec:patch_deletion}.

\par
To effectively incorporate Shapley values and interactions,
our approach adopts a greedy strategy based on the following self-context values. 
\begin{definition}\label{def:self_context}
    Let $N$ be a set of all players. 
    The \textbf{self-context Shapley value} of $S \subset N$ is defined by\footnote{While the Shapley value in Eq.~\eqref{eq:shapley} is defined for a single player, 
    we adapt it here to handle multiple players.}
    \begin{align}
        \phi^{\mathrm{sc}}(S) \stackrel{\mathrm{def}}{=} \phi(S|\{S\}) =  f(S)-f(\emptyset),
        \label{eq:self_context}
    \end{align}
     where the second equality follows from Eq~\eqref{eq:shapley}. 
\end{definition}
The self-context values were introduced in~\cite{moxi} as a conceptual analogy of Shapley values and interactions. 
In this work,
we reveal the connection between our framework and the original concepts.
We also introduce the $\pi^*$-index, which provides a perspective from sequential coalitional games and show that this index satisfies the essential axioms of coalitional games (see Sec.~\ref{sec:analysis_method} and App.~\ref{sec:axiom}).
Next, we define self-context interaction based on Eq.~\eqref{eq:interaction},
with a slight generalization to handle two or more players:
\begin{definition}
    The \textbf{self-context interaction} of $S \subset N$ is defined by
    \begin{align}
    I^{\mathrm{sc}}(S) \stackrel{\mathrm{def}}{=} \phi^{\mathrm{sc}}(S)-
    \sum_{S' \in \mathcal{P}^*(S)}\phi^{\mathrm{sc}}(S'),
    \label{eq:def_multi_interaction}
    \end{align}
    where $\mathcal{P}^*(S)$ denotes the set of all strict subsets of $S$, i.e., $ \mathcal{P}^*(S) = \mathcal{P}(S) \setminus \{S\}$.
\end{definition}
As shown in Eq.~\eqref{eq:interaction}, 
interaction is calculated by subtracting the individual contributions of each player from the joint contribution;
therefore, Eq.~\eqref{eq:def_multi_interaction} serves as a natural extension.

\par
Our greedy strategy in the insertion setup in Problem~\ref{problem:problem_insertion} works as follows. 
At each step $k = 1, \ldots, |N|/r$,\footnote{For simplicity, 
we assume that $r$ divides $|N|$.} 
a set $B_k$ of $r$ image patches is selected to greedily maximize the reward $f(B_1\cup \cdots \cup B_k)$. 
The set $B_k$ is selected by exhaustive reward evaluation of size-$r$ subsets of $N \setminus \bigcup_{i=1}^{k-1}B_{i}$.
The sum of (self-context) Shapley value and interaction naturally arises in this process, which justifies the consideration of interaction to address Problem~\ref{problem:problem_insertion}. 

\vspace{0.5em}
\noindent\textbf{Initial step (k=1).}
The set of patches $B_1 \subset N$ that maximizes $f(B_1)$  is selected. 
\begin{align}\label{eq:argmax_k1}
B_{1}  = \underset{B \subseteq N,|B|=r } {\operatorname{arg\,max}}\; f(B) - f(\emptyset)  = \underset{B \subseteq N,|B|=r  } {\operatorname{arg\,max}}\;  \phi^{\mathrm{sc}}(B).
\end{align}
The second equality derives from Eq.~\eqref{eq:self_context} and the fact that $f(\emptyset)$ is constant.
Note that we treat $B_1$ as a single player in the sense of applying the definition in Eq.~\eqref{eq:self_context}.
While this seems to consider only sets of size $r$, 
Eq.~\eqref{eq:def_multi_interaction} shows that this is not the case. 
From Eq~\eqref{eq:def_multi_interaction}, 
we have the following decomposition.
\begin{align}\label{eq:argmax_modify}
\phi^{\mathrm{sc}}(B)  =  I^{\mathrm{sc}}(B) + \sum_{B' \in \mathcal{P}^*(B)}\phi^{\mathrm{sc}}(B').
\end{align}
The first term of Eq.~\eqref{eq:argmax_modify} is the interaction among $r$ patches in $B$.
The second term is the sum of the self-context Shapley values over all subsets of $B$ except $B$ itself. 
Hence, the selection of $B_1$ takes into account various sizes of subsets and interactions of patches
(for $r=1$, only the individual contribution of $B$ is considered).

\vspace{0.5em}
\noindent\textbf{General k.}
The new set $B_k$ is determined in combination with the sets $\mathcal{B}_{k-1} = \bigcup_{i=1}^{k-1} B_i$ from the previous steps. 
As in the initial step, we have
\begin{align}\label{eq:argmax_k2}
    B_k & =\underset{B \subseteq N\setminus \mathcal{B}_{k-1},|B|=r} {\operatorname{arg\,max}} f(\{ \mathcal{B}_{k-1}\} \cup B) \\
    &  = \underset{B \subseteq N\setminus \mathcal{B}_{k-1},|B|=r}  {\operatorname{arg\,max}} \phi^{\mathrm{sc}}(\{\mathcal{B}_{k-1}\} \cup B).
\end{align}
Again, $ \phi^{\mathrm{sc}}(\{\mathcal{B}_{k-1}\} \cup B)$ can be interpreted as a sum of interaction and Shapley values. 
\begin{align}\label{eq:argmax_k2_modify}
    \phi^{\mathrm{sc}}(\{\mathcal{B}_{k-1}\} \cup B)  
    & = I^{\mathrm{sc}}(\{\mathcal{B}_{k-1}\} \cup B) \\
    & \quad \quad + \sum_{B' \in \mathcal{P}^*(\{\mathcal{B}_{k-1}\}\cup B)}\phi^{\mathrm{sc}}(B'),
\end{align}
where $\mathcal{B}_{k-1}$ is regarded as a single player.
Hence, 
the selection of $B_k$ considers the interaction between $r+1$ patches ($r$ patches in $B_k$ and a single patch treated as $\mathcal{B}_{k-1}$), 
as well as the Shapley values of each patch combination.
Importantly, these selections aim to maximize the sum of the interaction and Shapley values,
leading to the selection of informative patches without redundancy,
as defined in Def.~\ref{def:interaction}.

\begin{table*}[ht]
\caption{Results of AUC in insertion (Ins), deletion (Del), and overall (OA) metrics. Each result is derived from 1,000 detection instances with a predicted class score above $0.7$. The best scores are indicated in bold. }
\centering
\begin{tabular}{c|ccc|ccc||ccc|ccc}  \toprule
\multicolumn{1}{c}{Dataset} & \multicolumn{6}{c}{MS-COCO} & \multicolumn{6}{c}{PASCAL VOC} \\ \cmidrule{2-13}
\multicolumn{1}{c}{Object detector} & \multicolumn{3}{c}{DETR} & \multicolumn{3}{c}{Faster R-CNN} & \multicolumn{3}{c}{DETR} & \multicolumn{3}{c}{Faster R-CNN} \\  
\multicolumn{1}{c}{Metric} & \multicolumn{1}{c}{Ins $\uparrow$} & \multicolumn{1}{c}{Del $\downarrow$} & \multicolumn{1}{c}{OA$\uparrow$} & \multicolumn{1}{c}{Ins $\uparrow$} & \multicolumn{1}{c}{Del $\downarrow$}  & \multicolumn{1}{c}{OA $\uparrow$} & \multicolumn{1}{c}{Ins $\uparrow$} & \multicolumn{1}{c}{Del $\downarrow$}  & \multicolumn{1}{c}{OA $\uparrow$} & \multicolumn{1}{c}{Ins $\uparrow$} & \multicolumn{1}{c}{Del $\downarrow$}  & \multicolumn{1}{c}{OA  $\uparrow$} \\ \midrule

Grad-CAM   &.535 & .311 & .224 & .648 & .293 & .355 & .610 & .236 & .374 & .684 & .276 & .408\\

Grad-CAM++     & .794 & .154 & .640 & .860 & .166 & .694 & .766 & .185 & .581 & .823 & .188 & .624 \\

D-RISE    & .839 & .149 & .690 & .867 & .152 & .715 & .704 & .181 & .523 & .783 & .206 & .577 \\

SSGrad-CAM    & .595 & .225 & .370 & .827 & .160 & .667 & .660 & .170 & .483 & .862 & .143 & .719 \\

ODAM     & .659 & .150 & .509 & .854 & .148 & .706 & .581 & .199 & .383 & .849 & .144 & .705 \\

SSGrad-CAM++     & .871 & .114 & .757 & .900 & .126 & .774 & .813 & .166 & .647 & \textbf{.890} & .140 & .750 \\ \midrule

VX-CODE ($r=1$)      & .904 & .053 & .851 & .912 & .072 & .840 & \textbf{.841} & .073 & .768 & .827 & .067 & .760 \\

VX-CODE ($r=2$)      & .906 & \textbf{.052} & .854 & .918 & .069 & .849 & .835 &.069 & .766 & .841 & .066 & .775 \\

VX-CODE ($r=3$)      & \textbf{.909} & \textbf{.052} & \textbf{.857} & \textbf{.922} & \textbf{.067} & \textbf{.855} & .838 & \textbf{.067} & \textbf{.771} & .850 & \textbf{.063} & \textbf{.787} \\ \bottomrule
\end{tabular}
\label{tab:del_ins}
\end{table*}

\begin{figure*}[ht]
    \centering
    \includegraphics[scale=0.42]{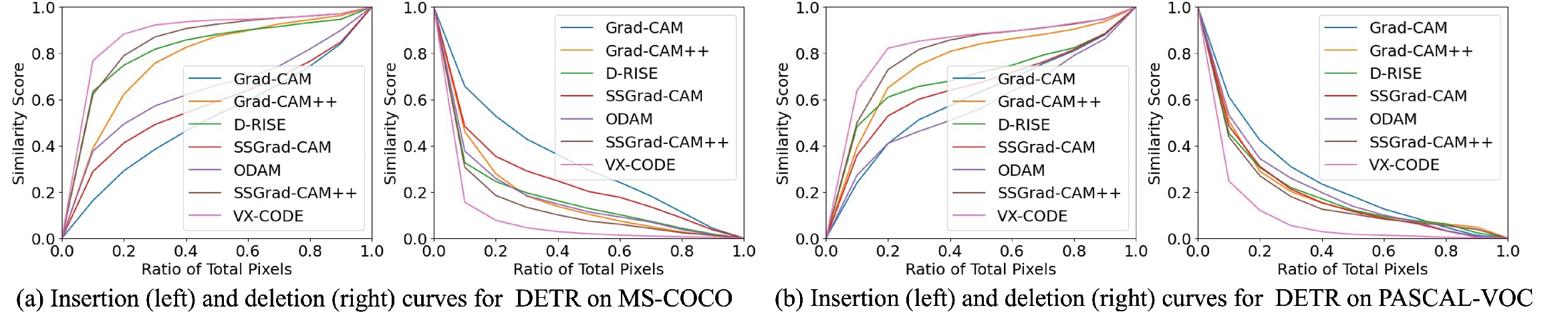}
    \caption{The insertion and deletion curves for DETR on (a) MS-COCO and (b) PASCAL VOC.}
    \label{fig:ins_del_curve}
\end{figure*}

\subsection{Further reducing computational cost}
\label{sec:reduction_patch_comb}
When $r{\ge}2$, 
the number of combinations ${}_{|N\setminus \mathcal{B}_{k-1}|}C_r$ increases combinatorially, 
making real-time execution challenging.
We therefore introduce two modules:
patch selection and step restriction.
The first limits the patches involved in the combination process, 
and the second confines the combination process to the early steps.
In \cref{sec:analysis_patch_combination}, 
we show that these modules significantly reduce generation time with only a slight decrease in accuracy.
See App.~\ref{sec:reduction_patch_comb_app} for details.

\par
\noindent\textbf{Patch selection.}\;
At each step, 
instead of considering all remaining patches, 
we restrict the combination candidates to the top-$m$
patches ranked by their single-patch reward impact.
As a result,
the number of combinations considered is reduced from ${}_{|N\setminus \mathcal{B}_{k-1}|}C_r$ to ${}_{m}C_r$. 
In our experiments, we set $m=30$.

\noindent\textbf{Step restriction.}\;
We limit the steps of the combination process by
stopping once $\gamma n$ ($0 \leq \gamma \leq 1$) patches have been identified,
and the remaining patches are then selected individually ($r=1$), which reduces computation.
Since the most important patches are usually identified in the early steps (see \cref{sec:analysis_patch_combination}),
this strategy reduces computation while preserving key-patch identification.
In our experiments, we set $\gamma=0.1$.

\subsection{Analysis}
\label{sec:analysis_method}
\paragraph{Sequential coalitional game.}
The self-context Shapley values and interactions are inspired by the original indices, and do not satisfy some axioms of coalitional games, e.g., efficiency. Although this does not affect the practical utility, we introduce the $\pi^*$-index, which satisfies the axioms (cf.~App.~\ref{sec:axiom}) and provides a perspective on our strategy as a sequential coalitional game. Recall $N = \{1, \ldots, n\}$ is the set of player IDs. Let $\Pi$ be the set of all permutations over $N$, where $\pi \in \Pi$ permutes a list $[1,\ldots, n]$ to $[\pi(1),\ldots, \pi(n)]$.
\begin{definition}
    Let $\pi^*\in\Pi$ be a permutation such that
    \begin{align} \label{eq:def_pi_star}
        \pi^* = \underset{\pi \in \Pi} {\operatorname{arg\,max}} \sum_{k=1}^{n} f(\pi(1), \ldots, \pi(k)).
    \end{align}
    The $\bm{\pi}^*$\textbf{-index} of player $k$ is defined by
    \begin{align}
        \psi(k | \pi^*) & \stackrel{\mathrm{def}}{=}  
         f(\pi^*(1),\ldots,\pi^*(k)) \\
         & \quad \quad \quad -f(\pi^*(1),\ldots,\pi^*(k-1)).
    \label{eq:def_ci}
    \end{align}
\end{definition}
We consider a sequential coalitional game where players are added in order, and the \piindex measures the contribution of each player by the reward gain upon their recruitment. The order $\pi^*$ maximizes the total reward by recruiting important players first. It is worth noting that \piindex focuses on the best case, whereas Shapley values correspond to the average case; 
indeed, averaging out $\psi(k \mid \pi)$ over all $\pi \in \Pi$  yields the Shapley value.

The proposed strategy in Sec.~\ref{sec:patch_insertion} can be viewed as a greedy approach to address the NP-hard computation of \piindex. From Eq.~\eqref{eq:argmax_k2}, we have
\begin{align}\label{eq:argmax_psi} &\psi(k| \hat{\pi}) = \phi^{\mathrm{sc}}(\{\mathcal{B}_{k-1}\} \cup B_{k}) - \phi^{\mathrm{sc}}(\mathcal{B}_{k-1}), \end{align} 
where $\hat{\pi}$ is the permutation obtained by the greedy selection (in hindsight) and $B_k$
is obtained by
\begin{align}
B_{k} & = \underset{B \subseteq N\setminus \mathcal{B}_{k-1},|B|=r} {\operatorname{arg\,max}} \phi^{\mathrm{sc}}(\{\mathcal{B}_{k-1}\} \cup B) \\ &= \underset{B \subseteq N\setminus \mathcal{B}_{k-1},|B|=r} {\operatorname{arg\,max}} \phi^{\mathrm{sc}}(\{\mathcal{B}_{k-1}\} \cup B) - \phi^{\mathrm{sc}}(\mathcal{B}_{k-1}).
\end{align}
As shown in Eq.~\eqref{eq:argmax_k2_modify}, the greedy selection considers interactions to determine $\hat{\pi}$. While we currently have no guarantee to obtain $\hat{\pi}=\pi^* $, our experiments demonstrate $\hat{\pi}$ already offer a strong benefit in identifying essential image patches to explain object detectors.

\begin{figure}[ht]
    \centering
    \includegraphics[scale=0.35]{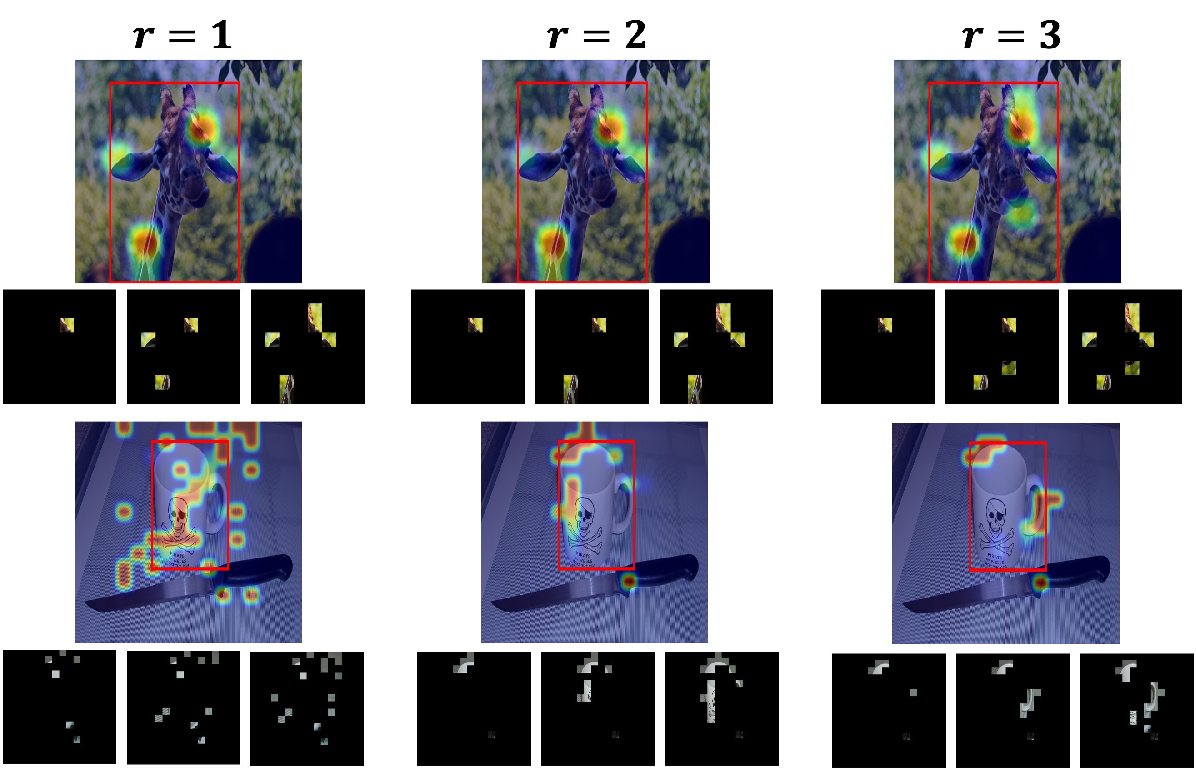}
    \caption{Comparison of heat maps generated by VX-CODE with patch insertion when changing $r$. Each heat map has some identified patches from left to right.}
    \label{fig:heatmap_a}
\end{figure}

\vspace{0.5em}
\noindent\textbf{Choice of $r$.}
Our framework considers both Shapley values and interactions,
and increasing $r$ enables evaluation over more patches (see App.~\ref{sec:patch_insertion_example} for examples with $r=1$ and $r=2$).
While $r=1$ already works well (see Sec.~\ref{sec:del_ins}), 
larger $r$ further improves the identification of important patches.
In greedy selection, 
each step affects the next; thus, 
failing to select important patches early can lead to an uninformative search. 
When $r=1$, 
limited early information may hinder the identification of important patches, 
and we observe that some cases fail as a result.
Increasing $r$ mitigates this by allowing consideration of various subset sizes and patch interactions (see Sec.~\ref{sec:compare_heatmap} and Fig.~\ref{fig:heatmap_a}).

\begin{figure*}[ht]
    \centering
    \includegraphics[scale=0.45]{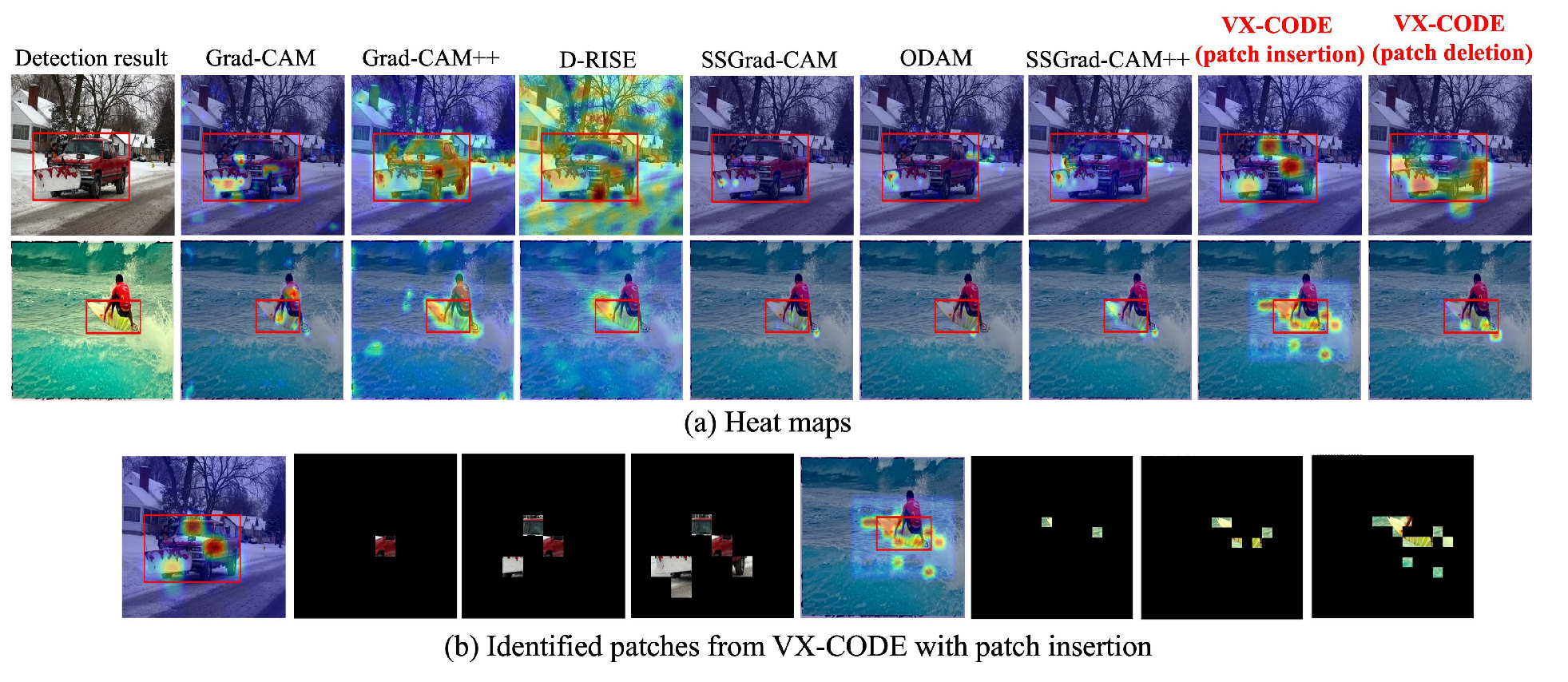}
    \caption{Comparison of visual explanations for DETR detections from each method. In VX-CODE, we use $r=1$. (a) Heat maps from each method. (b) Patches identified by VX-CODE using patch insertion.}
    \label{fig:heatmap}
\end{figure*}

\vspace{0.5em}
\noindent\textbf{Computational cost.}
Direct computation of Shapley values requires a cost of $\mathcal{O}(n\cdot 2^n)$. 
In contrast, our approach requires ${}_{|N\setminus \mathcal{B}_{k-1}|} C_r = \mathcal{O}(n^{r})$ reward evaluations for $n/r$ steps, resulting in a computational cost of $\mathcal{O}(n^{r+1}/r)$. 
While this is already tractable in practice, the cost grows with $r$.
For $r\geq2$,
the complexity can be reduced to $\mathcal{O}(\frac{\gamma n}{r}m^r + (1 - \gamma)n^2)$ via patch selection and step restriction.
In our experiments, we set $m = 30$ and $\gamma = 0.1$.

\section{Experiments}
\label{sec:experiments}
We compare VX-CODE with Grad-CAM~\cite{gradcam}, Grad-CAM++~\cite{gradcam++}, D-RISE~\cite{drise}, SSGrad-CAM~\cite{ssgcam}, ODAM~\cite{odam}, and SSGrad-CAM++~\cite{ssgcam++}.
For D-RISE, we use 5,000 masks at $16\times16$ resolution.
The same smoothing function is applied to SSGrad-CAM, ODAM, and SSGrad-CAM++.
For evaluation, we use MS-COCO~\cite{coco} and PASCAL VOC~\cite{voc} datasets,
which provide diverse scenes and object categories for robust comparison.
We apply each method to two object detectors: DETR~\cite{detr} and Faster R-CNN~\cite{faster_r_cnn}.\footnote{For PASCAL VOC, both models are trained with ImageNet pre-trained weights. For Faster R-CNN on MS-COCO, we use the pre-trained models from  \cite{detectron2}. For DETR on MS-COCO, we use the model from \url{https://github.com/facebookresearch/detr.git}.}
We use ResNet-50~\cite{resnet} as the backbone for DETR, and ResNet-50 and FPN~\cite{fpn} for Faster R-CNN.
All implementations are based on \textsc{detectron2}~\citep{detectron2}, 
a PyTorch-based library~\citep{pytorch}.
In VX-CODE, heat maps are generated from selected patches, as described in App.~\ref{sec:heatmap_generation}.
Implementation details are in App.~\ref{sec:implementation_details}.
To simulate patch removal in VX-CODE, 
we apply local zero masking~\cite{drise}.
We also evaluate blurring~\citep{blur1, SAG} as an alternative masking.
Additional results are provided in App.~\ref{sec:zero_blur}.

\subsection{Faithfulness of identified regions}
\label{sec:del_ins}
We evaluate faithfulness using the insertion and deletion metrics~\cite{RISE,drise},
which are widely used in visual explanation and enable fair comparison across methods (see App.~\ref{sec:limitation}). 
In insertion, each method adds important patches to a blank image.
In deletion, it removes them from the original image.
At each step, the methods run inference on a partially perturbed image and compute the similarity score to the original prediction.
Following~\cite{drise}, we define the similarity score using Eq.~\eqref{eq:reward}.
We compute the area under the curve (AUC) from these similarity scores.
Higher insertion and lower deletion AUCs indicate that important regions are identified early, 
demonstrating that the explanations are faithful.
We also report the overall metric~\cite{group-cam}, 
defined as the AUC difference between insertion and deletion.
In VX-CODE, insertion and deletion metrics are computed via patch insertion and deletion, respectively.

\par
Table~\ref{tab:del_ins} presents the average AUC for 1,000 detected instances with predicted class score above $0.7$.
VX-CODE with $r=1$ outperforms the existing methods on all metrics except insertion for Faster R-CNN on PASCAL VOC.
Increasing $r$ improves performance,
with $r=3$ achieving the best overall metric for all detectors and datasets.
Figure~\ref{fig:ins_del_curve} shows the insertion and deletion curves for DETR.
VX-CODE exhibits a steep rise in insertion
and a steep drop in deletion.
For example, on MS-COCO, 
VX-CODE surpasses the second-best method by 20\,\% in insertion and 50\,\% in deletion at the 10\,\% area mark.
This performance indicates that VX-CODE  identifies important regions using fewer pixels than the other methods.

\par
We evaluate VX-CODE on the one-stage detector RetinaNet~\cite{focal_loss} (App.~\ref{sec:one_stage}), 
compare it with Monte Carlo-based Shapley values (App.~\ref{sec:compare_shap}), 
and further analyze collective effects from the interaction perspective (App.~\ref{sec:collective_effects}).
We also evaluate class-level and box-level explanations separately (App.~\ref{sec:del_ins_class_score} and \ref{sec:box_explanation}), 
and assess localization performance using the pointing game~\cite{pointing_game} (App.~\ref{sec:pg}).

\subsection{Qualitative analysis of visualizations}
\label{sec:compare_heatmap}
Figure~\ref{fig:heatmap} shows explanations from each method for DETR detections.
In the first row of Fig.~\ref{fig:heatmap} (a), 
VX-CODE identifies multiple car features, 
including the body, windshield, and shovel, 
which jointly contribute to the detection. 
In contrast, the other methods either highlight irrelevant regions or focus only on the shovel.
In the second row  of Fig.~\ref{fig:heatmap} (a), 
VX-CODE with patch insertion captures both the surfboard and the surrounding sea surface, 
indicating that VX-CODE identifies background context as contributing to the detection.
In contrast, the other methods focus only on the surfboard and are less sensitive to the background.
VX-CODE with patch deletion focuses solely on the surfboard.
Patch deletion prioritizes removing instance features, 
which produce the largest drop in detection confidence.
As a result, the resulting explanations are object-centric and can improve
object localization.
We provide additional results in Apps.~\ref{sec:more_heatmap} and \ref{sec:heatmap_multi}, along with sanity checks (see App.~\ref{sec:sanity_checks}).

\par
Figure~\ref{fig:heatmap_a} demonstrates the effect of $r$, 
the number of patches inserted or deleted per step, on patch identification.
In the giraffe example, 
increasing $r$ yields broader feature coverage by accounting for more patch interactions:
$r=1$ and $r=2$ highlight the neck, forehead, and ears,
while $r=3$ also captures the nose.
In the mug example, larger $r$ values improve the accuracy of identifying important patches: 
at $r=1$,  critical patches (e.g., the mug's rim and handle) are missed, whereas $r=2$ and $r=3$ identify them earlier.
These results indicate that larger $r$ values better capture patch-level interactions by 
considering Shapley values and interactions among more patches.
Such a choice of larger $r$ helps reduce flat rewards, where reward increments are nearly identical and thus make it difficult to prioritize informative patches.
Additional examples are provided in App.~\ref{sec:more_patch_combination}.

\begin{table}[t]
\caption{Results of the deletion (Del) and insertion (Ins) for failure cases of mislocalization and misclassification from DETR on MS-COCO.
Each result is computed over 200 detection instances for each failure case.}
\centering
\begin{tabular}{c|cc|cc} \toprule
\multicolumn{1}{c}{Failure case} 
& \multicolumn{2}{c}{Misclassification}
& \multicolumn{2}{c}{Mislocalization} \\ \cmidrule{2-5}
\multicolumn{1}{c}{Metric} & Ins $\uparrow$ & Del $\downarrow$
& Ins $\uparrow$ & Del $\downarrow$   \\ \midrule

Grad-CAM 
& .463 & .286  
& .453 & .298  \\

Grad-CAM++
& .571 & .266   
& .643 & .200   \\

D-RISE
& .627 & .256  
& .759 & .133   \\

SSGrad-CAM
& .473 & .249 
& .501 & .263  \\

ODAM
& .413 & .243  
& .553 & .198  \\

SSGrad-CAM++
& .647 & .220  
& .766 & .120  \\ \midrule

Ours ($r=1$)
& .707 & .177
& .780 & .079 \\

Ours ($r=2$)
& .726 & .174 
& .782 & .079 \\

Ours ($r=3$)
& \textbf{.738} & \textbf{.168}
& \textbf{.787} & \textbf{.078} \\ \bottomrule

\end{tabular}
\label{tab:ins_del_failure_case}
\end{table}

\subsection{Quantitative results for failure cases}
\label{sec:failer_cases}
We quantitatively evaluate each method on two types of detection failures~\cite{erro_detector}: 
(1) misclassification, where an object is correctly localized but assigned an incorrect class label, 
and (2) mislocalization, where the object is detected with an inaccurate bounding box. 
Qualitative results for each case are shown in Fig.~\ref{fig:failer_case}, 
and additional examples are provided in App.~\ref{sec:eval_failer_case}.

Table~\ref{tab:ins_del_failure_case} summarizes the quantitative results, computed using the insertion and deletion metrics.
VX-CODE with $r=1$ already outperforms the existing methods on all metrics.
Increasing $r$ further enhances performance.
These results demonstrate that VX-CODE identifies 
failure-related regions more faithfully, as shown by consistently higher insertion and lower deletion AUCs.

These explanations also help reveal underlying biases in the model.
For example, bird training images often include sky backgrounds,  
which leads the model to misclassify images with sky, such as Fig.~\ref{fig:failer_case}, as birds.  Similarly, bicycle images often include riders,  
leading the model to rely on human presence for bicycle detection.
Beyond failure analysis,
we use VX-CODE to analyze behavioral differences between CNN and transformer-based models, 
with results presented in App.~\ref{sec:analysis_detector}.

\begin{table}[t]
\caption{Ablation study evaluating the effects of patch selection (PS) and step restriction (SR) with $r=2$. 
The check mark $\checkmark$ denotes the inclusion of each module. 
We present the deletion (Del), insertion (Ins), overall (OA), and explanation generation time per instance (Time), based on 50 instances detected by DETR on MS-COCO using an NVIDIA RTX A6000.}
\centering
\begin{tabular}{cc|cccc} \toprule
PS & SR & Ins $\uparrow$ & Del $\downarrow$ & OA $\uparrow$ & Time [s]$\downarrow$ \\ \midrule
 & & .917 & .057 & .860 & 768.126 \\
\checkmark & & .917 & .059 & .858& 287.610 \\
 & \checkmark & \textbf{.922} & \textbf{.056} & \textbf{.866} & 281.780 \\
\checkmark & \checkmark & .921 & .058 & .863 & \textbf{96.253} \\ \bottomrule
\label{tab:effect_a}
\end{tabular}
\end{table}

\subsection{Analysis of patch selection and step restriction}
\label{sec:analysis_patch_combination}
We reduce computational costs by applying patch selection and step restriction (see \cref{sec:reduction_patch_comb}) when $r\geq2$. 
Table~\ref{tab:effect_a} summarizes their effects with $r=2$.
As shown in Tab.~\ref{tab:effect_a}, 
generation time is substantial when patch selection and step restriction are not applied. 
In contrast, incorporating them significantly reduces generation time,
with only slight decreases in insertion and deletion metrics. 
Interestingly, applying step restriction improves all insertion–deletion metrics.
This observation suggests that informative patches are selected in the initial steps, 
making later-step combinations less beneficial.

\begin{table}[t]
  \centering
  \caption{Results of the deletion (Del), insertion (Ins), and overall (OA) using Grounding DINO.
  Each result is computed over 100 instances with the predicted class score above 0.6 on MS-COCO.}
  \label{tab:del_ins_dino}
  \begin{tabular}{c|ccc}
    \toprule
    Metric & Ins $\uparrow$ & Del $\downarrow$ & OA $\uparrow$ \\
    \midrule
    VPS & .946 & .300 & .646 \\
    VX-CODE (ours) & \textbf{.962} & \textbf{.211} & \textbf{.751} \\
    \bottomrule
  \end{tabular}
\end{table}

\subsection{Adaptation to object-level foundation model}
\label{sec:eval_dino}
VX-CODE is applicable to object-level foundation models such as Grounding DINO~\citep{grounding_dino}
and produces more faithful explanations.
We quantitatively compare VX-CODE with the recently introduced VPS~\citep{vps}.
Table~\ref{tab:del_ins_dino} presents the results by applying both methods to Grounding DINO~\citep{grounding_dino}.
We use VX-CODE with $r=1$.
VX-CODE outperforms VPS across all metrics, 
reflecting more faithful heat maps.
Additional comparisons with VPS are provided
in App.~\ref{sec:eval_vps}.

\section{Conclusion}
\label{sec:conclusion}
We propose VX-CODE, 
a visual explanation method for object detectors that considers the collective contributions of multiple pixels.
VX-CODE efficiently incorporates Shapley values and interactions over patch combinations to explain object detectors.
Furthermore, we introduce the \piindex and show its relationship to the sequential coalitional game.
We also describe techniques that reduce computational costs as the number of patch combinations increases.
Quantitative and qualitative experiments, including benchmark tests, bias analysis, and failure case studies, demonstrate that VX-CODE identifies important regions more faithfully than the baselines.
Overall, the results show that collective feature contributions are important for understanding how object detectors make predictions.

\section*{Acknowledgments}
This work was supported by JSPS KAKENHI Grant Number JP23K24914,
JST PRESTO Grant Number JPMJPR24K4,
JST BOOST Program Grant Number JPMJBY24C6,
and ROIS NII Open Collaborative Research 261S07-24168.

\clearpage
\setcounter{page}{1}
\maketitlesupplementary

\appendix
\renewcommand{\thesection}{\Alph{section}}
\renewcommand{\thefigure}{\Alph{figure}}

\section{Limitations and clarifications}
\label{sec:limitation}
\paragraph{Theoretical aspects of approximating accuracy in VX-CODE.}
Although the proposed method leverages Shapley values and interactions, theoretical analysis of the approximation accuracy is left for future work.
However,
we emphasize that the goal of this work is not to approximate Shapley
values—\textbf{it is to effectively identify important regions by considering collective contribution.}
Our method is inspired by and implements the idea of Shapley values and interactions in a computationally efficient manner, 
particularly via a greedy algorithm with a variant of the Shapley values
and interaction, 
as shown in our formulation (Sec.~\ref{sec:patch_insertion}).
Consequently, our method works well in model explanation.
We empirically evaluate the approximation accuracy by comparing it with Shapley values estimated via Monte Carlo sampling,
and the results are presented in App.~\ref{sec:compare_shap}.

\paragraph{Validity of reward function.}
To ensure a fair comparison, 
we use the same reward function Eq.~\eqref{eq:reward} used in existing studies, 
including those based on Shapley values~\cite{drise, bsed, fsod}. 
The design of an appropriate reward function is itself an open problem and is beyond the scope of this paper.
Importantly, 
the proposed method is agnostic to the choice of reward function and can be applied as-is once an appropriate one is defined in future work.

\paragraph{Validity of evaluation metrics.}
In this study, 
we primarily evaluate the proposed method using the insertion and deletion metrics.
These metrics have been long used in literature of visual explanation and feature selection,
and they enable fair comparison across methods.
Superior performance on these metrics is well aligned with providing convincing explanation of the models.
Note that our method outperforms the baselines not only in these metrics,
but also in another common metric, the pointing game (see App.~\ref{sec:pg}).
Additionally,
to demonstrate the effectiveness of the proposed method, 
we conduct extensive qualitative experiments. 
These include accurately identifying important regions outside the bounding box~(Sec.~\ref{sec:problem_setup} and App.~\ref{sec:more_bias_model}), 
introducing a weighted reward function and visualizing patches that contribute more to classification or localization~(App.~\ref{sec:change_reward}), 
and analyzing object detectors using the proposed method~(App.~\ref{sec:analysis_detector}). 
Such experiments have rarely been explored in the literature and represent one of the key contributions of this paper.

\section{Visual explanations for classification}
\label{sec:related_classification}
In the main paper, 
we omitted detailed discussions of related work on visual explanations for classification due to space constraints; here, we provide a more comprehensive review.

\par
Various visual explanation methods have been proposed for image classification models~\cite{Adebayo2018, LIME, gradcam, gradcam++, score_cam, RISE, x-gradcam, shap-cam, SAG, REX, moxi}. Generally, these methods can be broadly classified into three categories: gradient-based, class-activation-based (CAM-based), and perturbation-based.

\par
\textbf{Gradient-based methods.} \quad
These methods generate a saliency map by calculating the gradient of each pixel in the input image with respect to the model's output score, visualizing the important regions~\cite{Simonyan2013, Zeiler2013, Springenberg2015, relevance, smoothgrad, pointing_game, Sundararajan2017, deeplift, crp}.
For example, LRP~\cite{relevance} computes pixel-wise feature attributions by propagating relevance from the output layer to the input layer.
CRP~\cite{crp} extends LRP by emphasizing the relevance originating from the target class, calculating the disparity between the relevance from the target class and the average relevance from other classes.

\par
\textbf{CAM-based methods.} \quad
These methods calculate the weights representing the importance of each channel in the feature maps,
and visualize important regions by computing the weighted sum of these weights and the feature maps~\cite{cam, gradcam, gradcam++, score_cam, x-gradcam,ablation-cam, liftcam,shap-cam}.
CAM~\cite{cam} incorporates global average pooling into the CNN and generates a heat map that highlights important regions.
Grad-CAM~\cite{gradcam} generalizes CAM, making it possible to incorporate it into any CNN.
Grad-CAM++~\cite{gradcam++} improves upon Grad-CAM by incorporating pixel-level weighting, thereby enhancing the quality of the heat map.
Other methods that more accurately compute the weights representing the importance of each channel have been proposed~\cite{score_cam, x-gradcam, ablation-cam, shap-cam}.

\par
\textbf{Perturbation-based methods.} \quad
These methods identify important regions by introducing perturbations to the input image and quantifying the changes in the output~\cite{LIME,RISE,shap,many-shap,pred-diff,fast-shap, SAG, REX, moxi}.
LIME~\cite{LIME,slice} occludes by using super-pixel to compute the importance score.
RISE~\cite{RISE} visualizes important regions by applying random masks to the input image and analyzing the output results generated from the masked images.
SHAP~\cite{shap} distributes confidence scores to contributions by utilizing Shapley values from game theory.

\par
Methods that consider interactions~\cite{interaction} in addition to Shapley values have also been proposed~\cite{pred-diff,moxi}.
PredDIFF~\cite{pred-diff} is a method that measures changes in predictions by marginalizing features from a probabilistic perspective.
This paper clarifies the relationship between PredDIFF and Shapley values,
and introduces a new measurement method that takes interactions into account.
MoXI~\cite{moxi} utilizes a technique to approximate Shapley values and interactions to identify important regions for classification models.
While MoXI considers only pairwise patch interactions and their Shapley values 
(equivalent to $r=1$ in our method), 
our method considers interactions with arbitrary patches and their Shapley values.
As described in Sec.~\ref{sec:compare_heatmap}, 
in explanations for object detectors, 
some cases fail to identify important patches when using $r=1$.
The proposed method addresses this limitation and provides more faithful explanations for object detectors.
Additionally,
we introduce \piindex that clarifies the connection between its underlying concept and sequential coalitional games (see Sec.~\ref{sec:analysis_method}),
and prove that this index satisfies the essential axioms of coalitional games (see App.~\ref{sec:axiom}).

\par
Methods such as SAG~\cite{SAG} and REX~\cite{REX} have been proposed to generate multiple explanations for image classification models.
These methods determine importance rankings for each patch or pixel from multiple perspectives. 
For example, SAG~\cite{SAG} divides an image into $7\times7$ patches and generates multiple explanations by identifying the minimal sufficient explanation using beam search.
A similar approach to SAG and REX is the Visual Precision Search~\cite{vps}, which has been proposed for object foundation models to greedily identify important patches.
These methods are essentially similar to the patch insertion in the proposed method.
The key difference in the proposed method is that we highlight how these patch selections incorporate Shapley values and interactions and analyze object detectors from the perspective of collective contributions across multiple patches.
Additionally, the proposed method incorporates not only patch insertion but also patch deletion.

\par
Finally, we also mention logic-based explanations~\cite{logic1,logic2,logic3}.
These methods aim to produce highly reliable explanations, 
but they treat the model as a white box and represent it using logical formulas.
In contrast, the proposed method treats the model as a black box and does not rely on its internal structure.

\begin{algorithm}[t]
\caption{Process of patch insertion}
\label{alg:patch_insertion}
\begin{algorithmic}[1]
\REQUIRE Reward function $f$,\\
index set $N$ of patches, \\
the number of patches $r \in \{1, \ldots, |N|\}$ inserted at each step, \\
patch selection parameter $m \in \{1, \ldots, |N|\}$, \\
step restriction parameter $\gamma \in [0,1]$.
\ENSURE Sequence of subsets $\mathcal{B}_1, \ldots, \mathcal{B}_{E} \subseteq N$
\STATE $\mathcal{B}_k \leftarrow \{\}$ for all $k=0,\ldots, E$
\STATE $T=N$, $k=1$

\WHILE{$|T| \neq 0$}

\IF{$|T| < r$}
\STATE $r \leftarrow |T|$
\ENDIF

\IF{$r \geq 2$ and $|\mathcal{B}_{k-1}| \leq \gamma|N|$}
\STATE \texttt{\# Require} $|N \setminus \mathcal{B}_{k-1}|$ \texttt{forward process.}
\STATE $P^{m}_{k} \leftarrow \underset{P \subseteq N \setminus \mathcal{B}_{k-1}, |P|=m}{\operatorname{arg\,max}}\; \sum_{b \in P}f(\{\mathcal{B}_{k-1}\} \cup \{b\})$

\STATE \texttt{\# Require} ${}_{m} C_r$ \texttt{forward process.}
\STATE $B_{k} \leftarrow \underset{B \subseteq P^{m}_{k}, |B|=r}{\operatorname{arg\,max}}\; f(\{\mathcal{B}_{k-1}\} \cup B)$

\STATE $\mathcal{B}_k \leftarrow \mathcal{B}_{k-1} \cup B_k$
\STATE $T \leftarrow T \setminus B_k$
\STATE $k \leftarrow k+1$

\ELSE
\STATE \texttt{\# Require} $|N \setminus \mathcal{B}_{k-1}|$ \texttt{forward process.}
\STATE $b_k \leftarrow \underset{b \in N \setminus \mathcal{B}_{k-1}}{\operatorname{arg\,max}}\; f(\{\mathcal{B}_{k-1}\} \cup \{b\})$

\STATE $\mathcal{B}_k \leftarrow \mathcal{B}_{k-1} \cup \{b_k\}$
\STATE $T \leftarrow T \setminus \{b_k\}$
\STATE $k \leftarrow k+1$
\ENDIF
\ENDWHILE
\RETURN $\mathcal{B}_1, \ldots, \mathcal{B}_E$
\end{algorithmic}
\end{algorithm}

\begin{algorithm}[t]
\caption{Process of patch deletion}
\label{alg:patch_deletion}
\begin{algorithmic}[1]
\REQUIRE Reward function $f$, \\
index set $N$ of patches, \\
the number of patches $r \in \{1, \ldots, |N|\}$ removed at each step, \\
patch selection parameter $m \in \{1, \ldots, |N|\}$,\\
step restriction parameter $\gamma \in [0,1]$.
\ENSURE Sequence of subsets $\mathcal{B}_1, \ldots, \mathcal{B}_{E} \subseteq N$
\STATE $\mathcal{B}_k \leftarrow \{\}$ for all $k=0,\ldots, E$
\STATE $T=N$, $k=1$

\WHILE{$|T| \neq 0$}

\IF{$|T| < r$}
\STATE $r \leftarrow |T|$
\ENDIF

\IF{$r \geq 2$ and $|\mathcal{B}_{k-1}| \leq \gamma|N|$}
\STATE \texttt{\# Require} $|N \setminus \mathcal{B}_{k-1}|$ \texttt{forward process.}
\STATE $P^{m}_{k} \leftarrow \underset{P \subseteq N \setminus \mathcal{B}_{k-1},|P|=m}{\operatorname{arg\,max}}\; \sum_{b \in P}-f(N \setminus (\{\mathcal{B}_{k-1}\} \cup \{b\}))$

\STATE \texttt{\# Require} ${}_{m} C_r$ \texttt{forward process.}
\STATE $B_{k} \leftarrow \underset{B \subseteq P^{m}_{k}, |B|=r}{\operatorname{arg\,min}}\; f(N \setminus (\{\mathcal{B}_{k-1}\} \cup B))$

\STATE $\mathcal{B}_k \leftarrow \mathcal{B}_{k-1} \cup B_k$
\STATE $T \leftarrow T \setminus B_k$
\STATE $k \leftarrow k+1$

\ELSE
\STATE \texttt{\# Require} $|N \setminus \mathcal{B}_{k-1}|$ \texttt{forward process.}
\STATE $b_k \leftarrow \underset{b \in N \setminus \mathcal{B}_{k-1}}{\operatorname{arg\,min}}\; f(N \setminus (\{\mathcal{B}_{k-1}\} \cup \{b\}))$

\STATE $\mathcal{B}_k \leftarrow \mathcal{B}_{k-1} \cup \{b_k\}$
\STATE $T \leftarrow T \setminus \{b_k\}$
\STATE $k \leftarrow k+1$
\ENDIF
\ENDWHILE
\RETURN $\mathcal{B}_1, \ldots, \mathcal{B}_E$
\end{algorithmic}
\end{algorithm}

\section{Details of the proposed method}
\label{sec:detailed_method}
In this section,
we provide details of the proposed method that are not included in the main paper.

\subsection{Strategies for further reducing computational cost}
\label{sec:reduction_patch_comb_app}
When $r \geq 2$,
the proposed method leads to an increase in the number of combinations ${}_{|N\setminus \mathcal{B}_{k-1}|} C_r$, making real-time execution challenging.
To mitigate this, 
we introduce patch selection and step restriction.
Specifically, patch selection reduces the number of patches involved in the combination process to below $|N\setminus \mathcal{B}_{k-1}|$, 
and step restriction confines the combination process to the early steps.
While these methods reduce the total number of combinations and lower computational cost, they may lead to a decrease in accuracy.
In \cref{sec:analysis_patch_combination}, 
we show that these modules significantly reduce generation time with only a slight decrease in accuracy.

\paragraph{Patch selection.}
We limit the number of patches to be combined, assuming that combining each patch that has little impact on the reward is meaningless.
Let $m$ be the number of patches selected for combination,
and $P^{m}_{k}$ be the set of target patches to be combined at step $k$, where $|P^{m}_{k}|=m$.
Then, $P^{m}_{k}$ is determined as follows:
\begin{align}
P^{m}_{k}  = \underset{P \subseteq N \setminus \mathcal{B}_{k-1}}{\operatorname{arg\,max}}\; \sum_{b \in P}g(b), \quad  \text{s.t.}\ \ |P|=m, 
\label{eq:topl}
\end{align}
where 
\begin{align}
g(b) =
\begin{cases}
f(\{\mathcal{B}_{k-1}\} \cup \{b\}) & \text{(insertion)} \\
-f\left(N \setminus \left(\{\mathcal{B}_{k-1}\} \cup \{b\}\right)\right) & \text{(deletion)}
\label{eq:topl_eq}
\end{cases}.
\end{align}
Equation~\eqref{eq:topl} selects the set of patches with the top $m$ rewards.
This module efficiently selects target patches to maximize the reward function through combination.

\paragraph{Step restriction.}
We limit the steps of the combination process by
stopping once $\gamma n$ ($0 \leq \gamma \leq 1$) patches have been identified,
and the remaining patches are selected individually ($r=1$).
In our framework, the most important patches are typically identified in the initial steps (see App.~\ref{sec:analysis_patch_combination}).
Therefore, restricting the combination process to these steps allows efficient identification of important patches.

\paragraph{}
These two methods reduce the computational cost from $\mathcal{O}(r^{-1}n^{r+1})$ to $\mathcal{O}(\frac{\gamma n}{r}m^{r} + (1-\gamma)n^2)$.
We set $m=30$ and $\gamma=0.1$.

\subsection{Details of the algorithm}
\label{sec:pseudo_code}
We summarize the algorithm of VX-CODE.
The processes of patch insertion and patch deletion are shown in Algorithm~\ref{alg:patch_insertion} and Algorithm~\ref{alg:patch_deletion}, respectively.
As shown in these algorithms,
VX-CODE identifies $r$ patches or a single patch at each step based on the value of the reward function, using a simple greedy algorithm.

\begin{figure}[t]
    \centering
    \includegraphics[width=\linewidth]{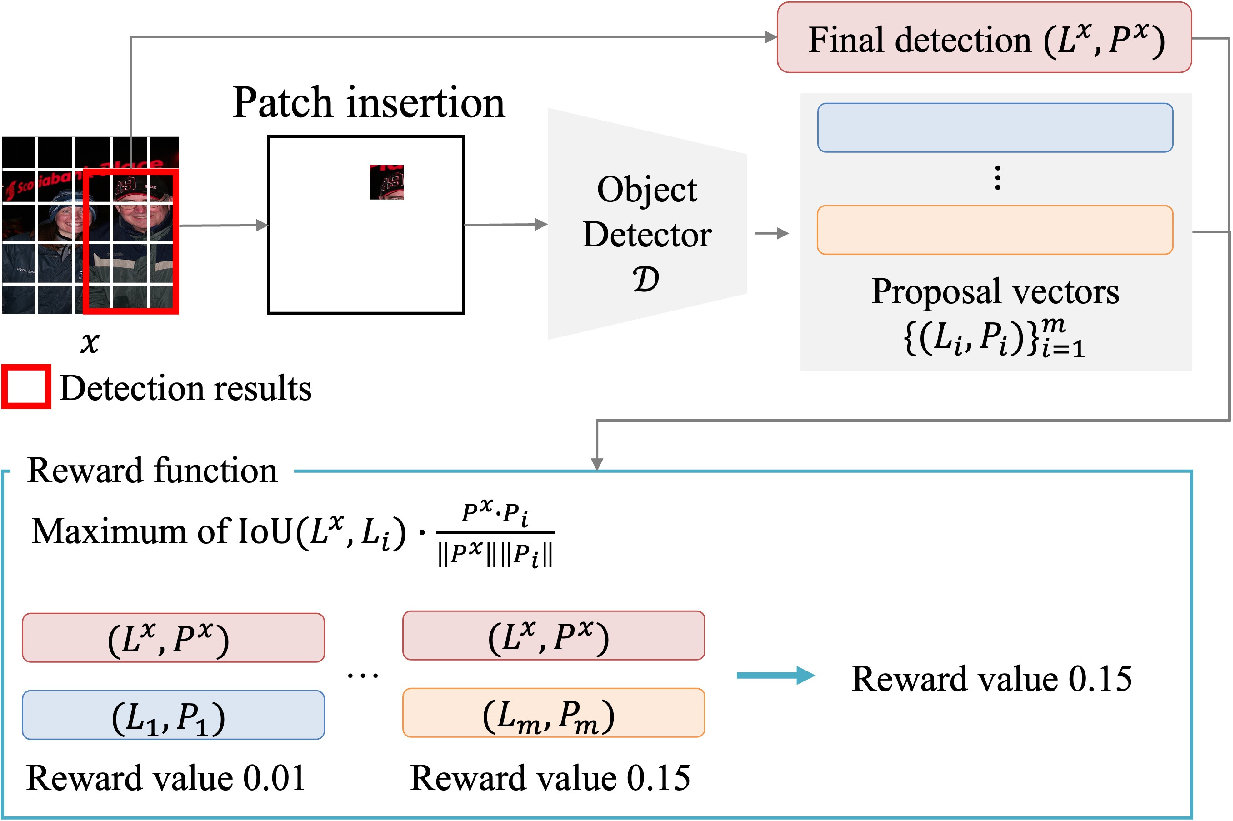}
    \caption{Conceptual illustration of reward computation.
    Given a partially corrupted image, the detector outputs multiple proposals.
    The reward is computed by scoring each proposal against the target detection in terms of localization and classification agreement, and taking the maximum score.}
    \label{fig:reward_concept}
\end{figure}

\subsection{Conceptual illustration of reward computation}
\label{sec:reward_concept}
Figure~\ref{fig:reward_concept} illustrates the concept of the reward computation in Eq.~\eqref{eq:reward}.
Given a partially corrupted image $x_S$, 
the detector outputs multiple proposals.
For each proposal, the reward evaluates how well it matches the target detection $(L^x, P^x)$ in terms of both localization and classification agreement.
The final reward is defined as the maximum score over all proposals.

\subsection{Specific example of patch insertion}
\label{sec:patch_insertion_example}
We provide specific examples for cases with $r=1$ and $r=2$ in patch insertion described in Sec.~\ref{sec:patch_insertion}.

\subsubsection{Case with r=1}
\label{sec:ins_case_1}
For step $k=1$, from Eqs.~\eqref{eq:argmax_k1} and \eqref{eq:argmax_modify},
only the highest Shapley value $\phi(\{b_{1}\}|\{\{b_{1}\}\})$ becomes the optimal set $\mathcal{B}_{1}=\{b_1\}$.

\par
For $k\geq2$, we find $b_k$ that maximizes $f(\{\mathcal{B}_{k-1}\} \cup \{b_k\})$.
From Eqs.~\eqref{eq:argmax_k2} and \eqref{eq:argmax_k2_modify}, this is formulated as follows:
\begin{align}
    b_k &=\underset{b \in N\setminus {\mathcal{B}_{k-1}}} {\operatorname{arg\,max}}\; f(\{\mathcal{B}_{k-1}\} \cup \{b\}) - f(\emptyset) \\        
    & = \underset{b \in N\setminus {\mathcal{B}_{k-1}}} {\operatorname{arg\,max}}\; \phi^{\mathrm{sc}}(\{\mathcal{B}_{k-1}\} \cup \{b\}) \\
    &= \underset{b \in N\setminus {\mathcal{B}_{k-1}}} {\operatorname{arg\,max}}\; I^{\mathrm{sc}}(\{\mathcal{B}_{k-1}\} \cup \{b\}) \\
    &\quad\quad\quad\quad + \sum_{B' \in \mathcal{P}^*(\{\mathcal{B}_{k-1}\}\cup \{b\})}\phi^{\mathrm{sc}}(B') \\    
    & = \underset{b \in N\setminus {\mathcal{B}_{k-1}}} {\operatorname{arg\,max}}\; I^{\mathrm{sc}}(\{\mathcal{B}_{k-1}\} \cup \{b\})+\phi^{\mathrm{sc}}(\{b\}) + \phi^{\mathrm{sc}}(\mathcal{B}_{k-1})  \\
    & = \underset{b \in N\setminus {\mathcal{B}_{k-1}}} {\operatorname{arg\,max}}\; I^{\mathrm{sc}}(\{\mathcal{B}_{k-1}\} \cup \{b\})+\phi^{\mathrm{sc}}(\{b\}).
\label{eq:argmax_k2_a1}
\end{align}
Therefore, this selection considers interactions between $\mathcal{B}_{k-1}$ and $b_k$,
as well as the Shapley values of $b_k$ in the self-context.

\subsubsection{Case with r=2}
For $k=1$,
we find $b_1$ and $b_2$ so that $f(\{b_1, b_2\})$ is maximized.
From Eqs.~\eqref{eq:argmax_k1} and \eqref{eq:argmax_modify}, this is formulated as follows: 
\begin{align}
    \{b_1, b_2\} &=\underset{\{b'_1, b'_2\} \subset N} {\operatorname{arg\,max}}\; f(\{b'_1, b'_2\}) - f(\emptyset) \\        
    & = \underset{\{b'_1, b'_2\} \subset N} {\operatorname{arg\,max}}\; \phi^{\mathrm{sc}}(\{b'_1, b'_2\}) \\
    &= \underset{\{b'_1, b'_2\} \subset N} {\operatorname{arg\,max}}\; I^{\mathrm{sc}}(\{b'_1, b'_2\})  \\
    &\quad\quad\quad\quad + \sum_{B' \in \mathcal{P}^*(\{b'_1, b'_2\})}\phi^{\mathrm{sc}}(B') \\
    & = \underset{\{b'_1, b'_2\} \subset N} {\operatorname{arg\,max}}\; I^{\mathrm{sc}}(\{b'_1, b'_2\}) \\
    & \quad\quad\quad\quad +\phi^{\mathrm{sc}}(\{b'_1\}) + \phi^{\mathrm{sc}}(\{b'_2\}).
\label{eq:argmax_k1_a2}
\end{align}
Therefore, 
this selection considers interactions between $b_1$ and $b_2$,
as well as their Shapley values in the self-context.

\par
For $k\geq2$, we find $b_{2k-1}$ and $b_{2k}$ so that $f(\{\mathcal{B}_{k-1}\} \cup \{b_{2k-1}, b_{2k}\})$ is maximized.
From Eqs.~\eqref{eq:argmax_k2} and \eqref{eq:argmax_k2_modify}, 
this is formulated as follows:
\begin{align}
     & \{b_{2k-1}, b_{2k}\} \\
     &=\underset{\{b'_{2k-1}, b'_{2k}\} \subseteq N\setminus {\mathcal{B}_{k-1}}} {\operatorname{arg\,max}}\; f(\{\mathcal{B}_{k-1}\} \cup \{b'_{2k-1}, b'_{2k}\}) - f(\emptyset) \\        
    & = \underset{\{b'_{2k-1}, b'_{2k}\} \subseteq N \setminus {\mathcal{B}_{k-1}}} {\operatorname{arg\,max}} \phi^{\mathrm{sc}}(\{\mathcal{B}_{k-1}\} \cup \{b'_{2k-1}, b'_{2k}\})  \\
    &= \underset{\{b'_{2k-1}, b'_{2k}\} \subseteq N \setminus {\mathcal{B}_{k-1}}} {\operatorname{arg\,max}}\; I^{\mathrm{sc}}(\{\mathcal{B}_{k-1}\} \cup \{b'_{2k-1}, b'_{2k}\})  \\
    &\quad\quad\quad\quad + \sum_{B' \in \mathcal{P}^*(\{\mathcal{B}_{k-1}\}\cup \{b'_{2k-1}, b'_{2k}\})}\phi^{\mathrm{sc}}(B') \\
    & = \underset{\{b'_{2k-1}, b'_{2k}\} \subseteq N \setminus {\mathcal{B}_{k-1}}} {\operatorname{arg\,max}}\; I^{\mathrm{sc}}(\{\mathcal{B}_{k-1}\} \cup \{b'_{2k-1}, b'_{2k}\})  \\
    & \quad\quad\quad\quad + \phi^{\mathrm{sc}}(\{\mathcal{B}_{k-1}\} \cup \{b'_{2k-1}\}) \\
    & \quad\quad\quad\quad + \phi^{\mathrm{sc}}(\{\mathcal{B}_{k-1}\} \cup \{b'_{2k}\}) \\
    & \quad\quad\quad\quad + \phi^{\mathrm{sc}}(\{b'_{2k-1}\} \cup \{b'_{2k}\}) \\
    & \quad\quad\quad\quad + \phi^{\mathrm{sc}}(\{b'_{2k-1}\}) \\
    & \quad\quad\quad\quad + \phi^{\mathrm{sc}}(\{b'_{2k}\}) \\
    & \quad\quad\quad\quad + \phi^{\mathrm{sc}}(\mathcal{B}_{k-1}) \\
    & = \underset{\{b'_{2k-1}, b'_{2k}\} \subseteq N \setminus {\mathcal{B}_{k-1}}} {\operatorname{arg\,max}}\; I^{\mathrm{sc}}(\{\mathcal{B}_{k-1}\} \cup \{b'_{2k-1}, b'_{2k}\}) \\
    & \quad\quad\quad\quad + \phi^{\mathrm{sc}}(\{\mathcal{B}_{k-1}\} \cup \{b'_{2k-1}\}) \\
    & \quad\quad\quad\quad + \phi^{\mathrm{sc}}(\{\mathcal{B}_{k-1}\} \cup \{b'_{2k}\})  \\
    & \quad\quad\quad\quad + \phi^{\mathrm{sc}}(\{b'_{2k-1}\} \cup \{b'_{2k}\}) \\
    & \quad\quad\quad\quad + \phi^{\mathrm{sc}}(\{b'_{2k-1}\}) \\
    & \quad\quad\quad\quad + \phi^{\mathrm{sc}}(\{b'_{2k}\}).
\label{eq:argmax_k2_a2}
\end{align}
Therefore, 
this selection considers interactions among the three elements $\mathcal{B}_{k-1}$, $b_{2k-1}$, and $b_{2k}$,
as well as Shapley values derived from each patch combination in self-context.

\subsection{Patch deletion}
\label{sec:patch_deletion}
We explain the process of patch deletion.
In this process,
we utilize the Shapley value, which measures the contribution in the absence of a player.
\begin{align}
\phi_{\mathrm{d}}(i \mid N) \stackrel{\mathrm{def}}{=}  \sum_{D \subseteq N, i \in D} P_{\mathrm{d}}(D \setminus \{i\} \mid N)[f(D) - f(D \setminus \{i\})],
\label{eq:shapley_absence}
\end{align}
where $P_{\mathrm{d}}(A \mid B) = \frac{(|B| - |A|-1)!|A|!}{|B|!}$.
This Shapley value quantifies the average impact attributable to the removal of player $i$.
To effectively incorporate Shapley values and interactions,
our approach adopts a greedy strategy based on the
following full-context values.
\begin{definition}\label{def:full_context}
    Let $N$ be a set of all players. 
    The \textbf{full-context Shapley value} of $S \subset N$ is defined by
    \begin{align}
    \phi_{\mathrm{d}}^{\mathrm{fc}}(S \mid N) \stackrel{\mathrm{def}}{=} P_{\mathrm{d}}(N \setminus S \mid N)[f(N) - f(N \setminus S)].
    \label{eq:shapley_fc}
    \end{align}    
\end{definition}
The full-context values were introduced in~\cite{moxi}.
These values restrict $D$ in Eq.~\eqref{eq:shapley_absence} to a fixed set.
Next, we define full-context interaction as follows:
\begin{definition}
    The \textbf{full-context interaction} of $S \subset N$ is defined by
    \begin{align}
    & I_{\mathrm{d}}^{\mathrm{fc}}(S) \stackrel{\mathrm{def}}{=} 
    \phi_{\mathrm{d}}^{\mathrm{fc}}(S \mid (N \setminus S) \cup \{S\}) \\
    & \quad\quad\quad\quad - \sum_{S' \in \mathcal{P}^*(S)} \phi_{\mathrm{d}}^{\mathrm{fc}}(S' \mid N \setminus (S \setminus S')).
    \label{eq:def_multi_interaction_absence}
    \end{align}
    where $\mathcal{P}^*(S)$ denotes the set of all strict subsets of $S$, i.e., $ \mathcal{P}^*(S) = \mathcal{P}(S) \setminus \{S\}$.
\end{definition}

\par
Similar to patch insertion,
our greedy strategy in the deletion setup in Problem~\ref{problem:problem_insertion} works as follows.

\paragraph{Initial step (k=1).} The set of patches $B_1 \subset N$ that minimize $f(N \setminus B_1)$  is selected. 
\begin{align}
    B_1 & = \underset{B \subseteq N,|B|=r} {\operatorname{arg\,min}}\; f(N \setminus B) - f(N)   \\
    & = \underset{B \subseteq N,|B|=r} {\operatorname{arg\,max}}\; f(N) - f(N \setminus B)  \\
    & = \underset{B \subseteq N,|B|=r} {\operatorname{arg\,max}}\; \phi_{\mathrm{d}}^{\mathrm{fc}}(B \mid N),
\label{eq:argmin_k1}
\end{align}
In the last equality, we omit $P_{\mathrm{d}}(N \setminus B \mid N)$, as it remains constant regardless of the patch selection.
Note that we treat $B_1$ as a single player. 
While this apparently considers the Shapley value of a fixed set size $r$ only, 
we can show that this is not the case. 
From Eq~\eqref{eq:def_multi_interaction_absence}, 
we have the following decomposition.
\begin{align}
   \phi_{\mathrm{d}}^{\mathrm{fc}}(B \mid N) = I_{\mathrm{d}}^{\mathrm{fc}}(B) + \sum_{B' \in \mathcal{P}^*(B)} \phi_{\mathrm{d}}^{\mathrm{fc}}(B' \mid N \setminus (B \setminus B')).
\label{eq:argmin_k1_modify}
\end{align}
The first term of Eq.~\eqref{eq:argmin_k1_modify} is the interaction among $r$ patches in $B$.
The second term is the sum of the full-context Shapley values over all subsets of $B$ except $B$ itself. 
Hence, the selection of $B_1$ takes into account various size of subsets and interactions of patches
(for $r=1$, only the individual contribution of $B$ is considered).

\paragraph{General k.} The new set $B_k$ is determined in combination with the sets $\mathcal{B}_{k-1} = \bigcup_{i=1}^{k-1} B_i$ from the previous steps. 
As in the initial step, we have
\begin{align}
    B_{k} &= \underset{B \subseteq N \setminus \mathcal{B}_{k-1},|B|=r } {\operatorname{arg\,min}}\; 
    f(N \setminus (\{\mathcal{B}_{k-1}\} \cup B)) - f(N) \\
    &= \underset{B \subseteq N \setminus \mathcal{B}_{k-1},|B|=r} {\operatorname{arg\,max}}\; 
    f(N)-f(N \setminus (\{\mathcal{B}_{k-1}\} \cup B)) \\
    & = \underset{B \subseteq N \setminus \mathcal{B}_{k-1},|B|=r} {\operatorname{arg\,max}}\; 
    \phi_{\mathrm{d}}^{\mathrm{fc}}(\{\mathcal{B}_{k-1}\} \cup B \mid N).
\label{eq:argmin_k2}
\end{align}
Again, $\phi_{\mathrm{d}}^{\mathrm{fc}}(\{\mathcal{B}_{k-1}\} \cup B \mid N)$ can be interpreted as a sum of interaction and Shapley values. 
\begin{align}\label{eq:argmin_k2_modify}
    & \phi_{\mathrm{d}}^{\mathrm{fc}}(\{\mathcal{B}_{k-1}\} \cup B \mid N) \\
    & = I_{\mathrm{d}}^{\mathrm{fc}}(\{\mathcal{B}_{k-1}\} \cup B) \\ 
    & \quad \quad+ \sum_{B' \in \mathcal{P}^*(\{\mathcal{B}_{k-1}\} \cup B)} \phi_{\mathrm{d}}^{\mathrm{fc}}(B' \mid N \setminus ((\{\mathcal{B}_{k-1}\}\cup B) \setminus B')).  
\end{align}
where $\mathcal{B}_{k-1}$ is regarded as a single patch.
Hence, 
the selection of $B_k$ considers the interaction between $r+1$ patches ($r$ patches in $B_k$ and a single patch treated as $\mathcal{B}_{k-1}$), 
as well as the Shapley values of each patch combination.

\subsection{Specific example of patch deletion}
\label{sec:patch_deletion_example}
We provide specific examples for cases with $r=1$ and $r=2$ in patch deletion described in App.~\ref{sec:patch_deletion}.

\subsubsection{Case with r=1}
\label{sec:del_case_1}
For step $k=1$,
from Eqs.~\eqref{eq:argmin_k1} and \eqref{eq:argmin_k1_modify},  
only the highest full-context Shapley value $\phi_{\mathrm{d}}^{\mathrm{fc}}(\{b_1\}|N) = n^{-1}[f(N)-f(N\setminus \{b_1\})]$ becomes the optimal set $B_{1}=\{b_1\}$.

\par
For $k\geq2$, we find $b_k$ that minimizes $f(N \setminus \{\{\mathcal{B}_{k-1}\} \cup \{b_k\}\})$.
From Eqs.~\eqref{eq:argmin_k2} and \eqref{eq:argmin_k2_modify}, 
this is formulated as follows:
\begin{align}
    b_k &= \underset{b \in N \setminus \mathcal{B}_{k-1}} {\operatorname{arg\,min}}\; 
    f(N \setminus (\{\mathcal{B}_{k-1}\} \cup \{b\})) - f(N) \\
    &= \underset{b \in N \setminus \mathcal{B}_{k-1}} {\operatorname{arg\,max}}\; 
    f(N) - f(N \setminus (\{\mathcal{B}_{k-1}\} \cup \{b\})) \\
    &= \underset{b \in N \setminus \mathcal{B}_{k-1}} {\operatorname{arg\,max}}\; 
    \phi_{\mathrm{d}}^{\mathrm{fc}}(\{\mathcal{B}_{k-1}\} \cup \{ b\} \mid N) \\
    &= \underset{b \in N \setminus \mathcal{B}_{k-1}} {\operatorname{arg\,max}}\; 
    I_{\mathrm{d}}^{\mathrm{fc}}(\{\mathcal{B}_{k-1}\} \cup \{b\})  \\
    & \quad + \sum_{B' \in \mathcal{P}^*(\{\mathcal{B}_{k-1}\} \cup \{b\})} \phi_{\mathrm{d}}^{\mathrm{fc}}(B' \mid N \setminus ((\{\mathcal{B}_{k-1}\} \cup \{b\}) \setminus B')) \\
    &= \underset{b \in N \setminus \mathcal{B}_{k-1}} {\operatorname{arg\,max}}\; 
    I_{\mathrm{d}}^{\mathrm{fc}}(\{\mathcal{B}_{k-1}\} \cup \{b\})  \\
    & \quad \quad \quad + \phi_{\mathrm{d}}^{\mathrm{fc}}(\{b\} \mid N \setminus \mathcal{B}_{k-1}) + \phi_{\mathrm{d}}^{\mathrm{fc}}(\mathcal{B}_{k-1} \mid N \setminus \{b\}).
\label{eq:argmin_k2_a1}
\end{align}
Therefore, this selection considers interactions between $\mathcal{B}_{k-1}$ and $b_k$,
as well as Shapley values in the full-context.

\subsubsection{Case with r=2}
For $k=1$,
we find $b_1$ and $b_2$ so that $f(N \setminus \{b_1, b_2\})$ is minimized.
From Eqs.~\eqref{eq:argmin_k1} and \eqref{eq:argmin_k1_modify},
this is formulated as follows:
\begin{align}
    \{b_1, b_2\} &= \underset{\{b'_1, b'_2\} \subset N} {\operatorname{arg\,min}}\; 
    f(N \setminus \{b'_1, b'_2\}) - f(N) \\
    &= \underset{\{b'_1, b'_2\} \subset N} {\operatorname{arg\,max}}\; 
    f(N) - f(N \setminus \{b'_1, b'_2\}) \\   
    &= \underset{\{b'_1, b'_2\} \subset N} {\operatorname{arg\,max}}\; 
    \phi_{\mathrm{d}}^{\mathrm{fc}}(\{b'_1, b'_2\} \mid N) \\
    &= \underset{\{b'_1, b'_2\} \subset N} {\operatorname{arg\,max}}\; 
    I_{\mathrm{d}}^{\mathrm{fc}}(\{b'_1, b'_2\}) \\
    & \quad + \sum_{B' \in \mathcal{P}^*(\{b'_1, b'_2\})} \phi_{\mathrm{d}}^{\mathrm{fc}}(B' \mid N \setminus (\{b'_1, b'_2\}\setminus B')) \\
    &= \underset{\{b'_1, b'_2\} \subset N} {\operatorname{arg\,max}}\; 
    I_{\mathrm{d}}^{\mathrm{fc}}(\{b'_1, b'_2\}) \\
    & \quad + \phi_{\mathrm{d}}^{\mathrm{fc}}(\{b'_1\} \mid N\setminus \{b'_2\}) + \phi_{\mathrm{d}}^{\mathrm{fc}}(\{b'_2\} \mid N\setminus \{b'_1\}).
\label{eq:argmin_k1_a2}
\end{align}
Therefore, this selection considers interactions between $b_1$ and $b_2$,
as well as their Shapley values in the full-context.

\par
For $k\geq2$, we find $b_{2k-1}$ and $b_{2k}$ so that $f(N \setminus \{\{\mathcal{B}_{k-1}\} \cup \{b_{2k-1}, b_{2k}\}\})$ is minimized.
From Eqs.~\eqref{eq:argmin_k2} and \eqref{eq:argmin_k2_modify}, 
this is formulated as follows:
\begin{align}
    & \{b_{2k-1}, b_{2k}\} \\
    & = \underset{\{b'_{2k-1}, b'_{2k}\} \subseteq N \setminus \mathcal{B}_{k-1}} {\operatorname{arg\,min}}
    f(N \setminus (\{\mathcal{B}_{k-1}\} \cup \{b'_{2k-1}, b'_{2k}\}))  - f(N) \\
    & = \underset{\{b'_{2k-1}, b'_{2k}\} \subseteq N \setminus \mathcal{B}_{k-1}} {\operatorname{arg\,max}}\; 
    f(N) - f(N \setminus (\{\mathcal{B}_{k-1}\} \cup \{b'_{2k-1}, b'_{2k}\})) \\
    & = \underset{\{b'_{2k-1}, b'_{2k}\} \subseteq N \setminus \mathcal{B}_{k-1}} {\operatorname{arg\,max}}\; 
    \phi_{\mathrm{d}}^{\mathrm{fc}}(\{\mathcal{B}_{k-1}\} \cup \{b'_{2k-1}, b'_{2k}\} \mid N) \\
    & = \underset{\{b'_{2k-1}, b'_{2k}\} \subseteq N \setminus \mathcal{B}_{k-1}} {\operatorname{arg\,max}}\; 
    I_{\mathrm{d}}^{\mathrm{fc}}(\{\mathcal{B}_{k-1}\} \cup \{b'_{2k-1}, b'_{2k}\}) \\
    & 
    \quad + \sum_{B' \in \mathcal{P}^*(\{\mathcal{B}_{k-1}\} \cup \{b'_{2k-1}, b'_{2k}\})} \phi_{\mathrm{d}}^{\mathrm{fc}}(B' \mid N \setminus \\
    & \qquad\qquad\qquad\qquad\qquad\qquad 
    ((\{\mathcal{B}_{k-1}\} \cup \{b'_{2k-1}, b'_{2k}\}) \setminus B')) \\
    & = \underset{\{b'_{2k-1}, b'_{2k}\} \subseteq N \setminus \mathcal{B}_{k-1}} {\operatorname{arg\,max}}\; 
    I_{\mathrm{d}}^{\mathrm{fc}}(\{\mathcal{B}_{k-1}\} \cup \{b'_{2k-1}, b'_{2k}\})  \\
    & \quad\quad\quad\quad + \phi_{\mathrm{d}}^{\mathrm{fc}}(\{b'_{2k-1}\} \cup \{b'_{2k}\} \mid N\setminus \mathcal{B}_{k-1})  \\
    & \quad\quad\quad\quad + \phi_{\mathrm{d}}^{\mathrm{fc}}(\{\mathcal{B}_{k-1}\} \cup \{b'_{2k-1}\} \mid N\setminus \{b'_{2k}\}) \\
    & \quad\quad\quad\quad + \phi_{\mathrm{d}}^{\mathrm{fc}}(\{\mathcal{B}_{k-1}\} \cup \{b'_{2k}\} \mid N\setminus \{b'_{2k-1}\}) \\
    & \quad\quad\quad\quad + \phi_{\mathrm{d}}^{\mathrm{fc}}(\{b'_{2k-1}\} \mid N\setminus (\{\mathcal{B}_{k-1}\} \cup \{b'_{2k}\})) \\
    & \quad\quad\quad\quad + \phi_{\mathrm{d}}^{\mathrm{fc}}(\{b'_{2k}\} \mid N\setminus (\{\mathcal{B}_{k-1}\} \cup \{b'_{2k-1}\}))  \\
    & \quad\quad\quad\quad + \phi_{\mathrm{d}}^{\mathrm{fc}}(\mathcal{B}_{k-1} \mid  N\setminus (\{b'_{2k-1}\} \cup \{b'_{2k}\})).
\label{eq:argmin_k2_a2}
\end{align}
Therefore,
this selection considers interactions among the three elements $\mathcal{B}_{k-1}$, $b_{2k-1}$, and $b_{2k}$,
as well as Shapley values derived from each patch combination in full-context.

\begin{algorithm}[t]
\caption{Generation of a heat map based on identified patches}
\label{alg:heatmap_generation}
\begin{algorithmic}[1]
\REQUIRE Set of identified patches $\{b_1, \ldots, b_{n}\}$,\\
set of values from a reward function $\{f_1, \ldots, f_{n}\}$.
\ENSURE Heat map $M \in \mathbb{R}^{h \times w}$ for an input image of size $h\times w$.
\STATE $M = \mathbf{0}$
\FOR{$i=1, \ldots, n$}

\IF{patch insertion}
\IF{$i=1$}
\STATE $s \leftarrow 1.0$
\ELSE
\STATE $s \leftarrow 1.0 - f_{i-1}$
\ENDIF
\ELSIF{patch deletion}
\IF{$i=1$}
\STATE $s \leftarrow 1.0$
\ELSE
\STATE $s \leftarrow f_{i-1}$
\ENDIF
\ENDIF

\STATE \texttt{\# Identify the diagonal corners of the patch $b_i$ in the image.}
\STATE $\{(x_1^{\mathrm{b}}, y_1^{\mathrm{b}}), (x_2^{\mathrm{b}}, y_2^{\mathrm{b}})\} \leftarrow \mathrm{get\_position}(b_i)$
\STATE \begin{align}
        M_{i,j} \leftarrow \nonumber
        \begin{cases}
        s & \begin{aligned}[t] & (\text{if } y_1^{\mathrm{b}} \leq i \leq y_2^{\mathrm{b}} \\
        & \quad \text{and } x_1^{\mathrm{b}} \leq j \leq x_2^{\mathrm{b}}), 
        \end{aligned} \\
        M_{i,j} & (\text{otherwise}). \\
        \end{cases}
        \end{align}

\ENDFOR
\RETURN $M$
\end{algorithmic}
\end{algorithm}

\subsection{Heat maps generation from identified patches}
\label{sec:heatmap_generation}
We describe the process for generating heat maps from identified patches.
To efficiently identify important regions, 
it is desirable to emphasize patches that lead to significant changes in the reward function values, either through patch insertion or patch deletion.
To address this, 
given a set of identified patches, $\{b_1, \ldots, b_{n}\}$
and their corresponding reward function values for each step,
$\{f_1, \ldots, f_{n}\}$, where $f_k$ is the reward function value at step $k$,
we determine the importance of each patch based on the corresponding reward function value.
Specifically, 
the first identified patch, 
which contributes most significantly, 
is assigned the highest importance value of 1.0.
For the subsequent patches, we assume that they contribute to the remaining change in the score
(an increase for patch insertion or a decrease for patch deletion) and define their importance
directly from the reward value at the previous step.
We show the pseudo-code for the heat map generation process in Algorithm~\ref{alg:heatmap_generation}.
In Algorithm~\ref{alg:heatmap_generation}, $\mathrm{get\_position}(b_i)$ returns the diagonal corners $(x_1^{\mathrm{b}}, y_1^{\mathrm{b}})$ and $(x_2^{\mathrm{b}}, y_2^{\mathrm{b}})$ of the patch $b_i$ in the image. 
This process generates a heat map that effectively highlights regions with significant changes in the reward function values.

\par
In the above framework, 
for $r\geq2$, it is necessary to assign a reward function value individually to each patch in the identified patch set $B_{k}$.
Therefore, we individually assign this value by sequentially adding each patch in $B_{k}$ to $\mathcal{B}_{k-1}$ and computing the corresponding reward function value.
In this process, the order of adding patches is determined such that the sum of the corresponding rewards is maximized (patch insertion) or minimized (patch deletion).
Although this requires $r!$ computations for all possible arrangements of the $r$ patches,
the practical values of $r$ are small (e.g., $r\leq5$), 
making the computational cost negligible in the overall algorithm.

\subsection{Implementation details}
\label{sec:implementation_details}
We describe the method for patch decomposition. 
Unlike image classification models, 
object detectors detect instances of varying sizes. 
Consequently, the size of each feature (e.g., a human head, hand, etc.) depends on the instance size. 
Ideally, each divided patch should capture these features. 
Therefore, we determine the patch size based on the size of the detected bounding boxes. 
Specifically, the patch size is determined as follows:
\begin{align}
    d_\mathrm{h} = d_\mathrm{w} = 
    \begin{cases}
    24 & (0 < R_{\mathrm{box}} \leq 0.01)\\
    16 & (0.01 < R_{\mathrm{box}} \leq 0.2)\\
    8 & (0.2 < R_{\mathrm{box}})
    \end{cases},
\label{eq:divide}
\end{align}
where $d_\mathrm{h}$ and $d_\mathrm{w}$ represent the vertical and horizontal divisions of the image, respectively, 
and $R_{\mathrm{box}}$ denotes the ratio of the detected box area to the total image area.

\par
Next, we describe the patches considered for calculation in patch insertion and patch deletion. 
For patches divided according to Eq.~\eqref{eq:divide}, 
patch insertion and patch deletion are performed.
In this process, particularly when the detected instance size is small, 
it is assumed that features from regions far from the instance contribute little to the result (e.g., an object located in the lower right of the image is irrelevant to information in the upper left). 
Additionally, as the patch size becomes smaller, 
computing for all patches incurs a high computational cost. 
Therefore, for the detection results with $R_{\mathrm{box}} \leq 0.2$, 
we limit the target patches for reward computation.
Let $\{x_1, y_1, x_2, y_2\}$ be a set of coordinates of the detected bounding box, 
and let $h$ and $w$ be the height and width of the image.
Then, we compute the reward only for the patches with center coordinates $(C_x, C_y)$ within the following range.
\begin{align}
    & x_1 - \frac{w}{7} \leq  C_x \leq x_2 + \frac{w}{7} \nonumber \\
    & y_1 - \frac{h}{7} \leq  C_y \leq y_2 + \frac{h}{7}.
\end{align}

\begin{figure}[t]
    \centering
    \includegraphics[scale=0.25]{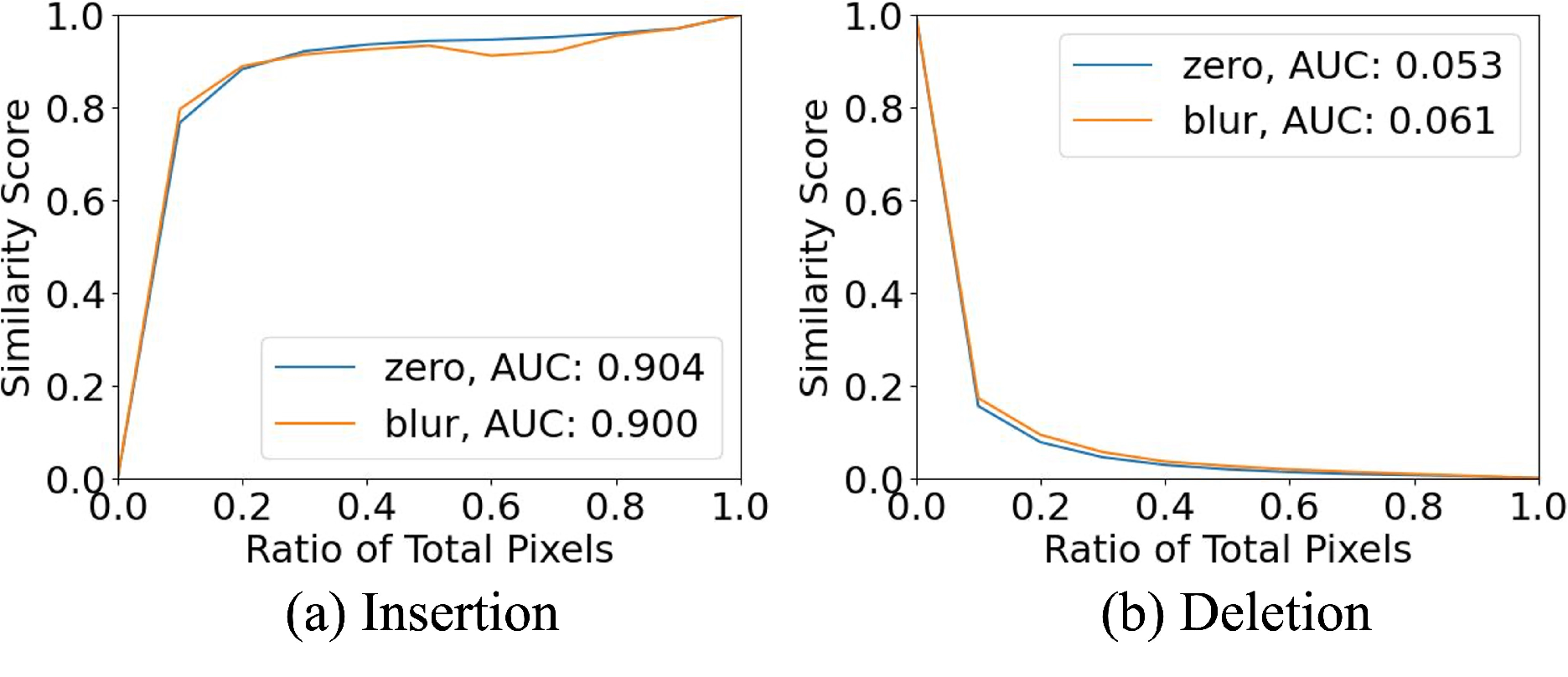}
    \caption{Comparison results of the two filling methods, zero filling and blurred pixel values, using (a) the insertion metric and (b) the deletion metric. The similarity score is computed using Eq.~\eqref{eq:reward}.}
    \label{fig:zero_blur}
\end{figure}

\begin{figure}[t]
    \centering
    \includegraphics[scale=0.5]{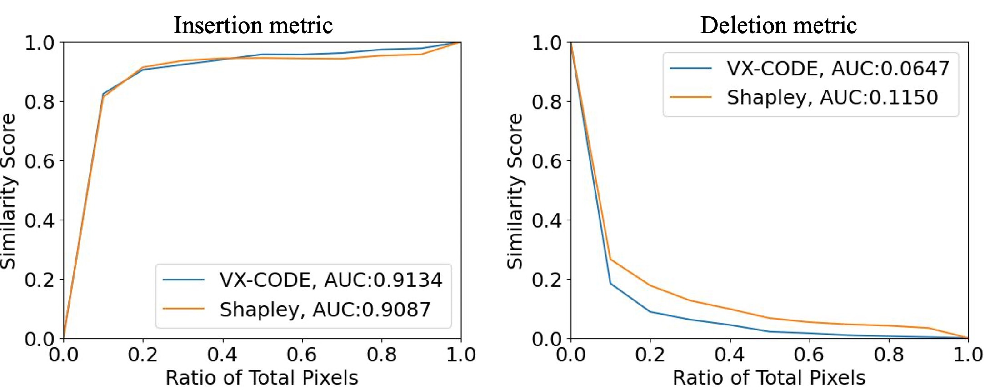}
    \caption{The comparison of insertion and deletion metrics between VX-CODE and Shapley values estimated via Monte Carlo sampling.}
    \label{fig:compare_shap}
\end{figure}

\begin{figure}[t]
    \centering
    \includegraphics[scale=0.7]{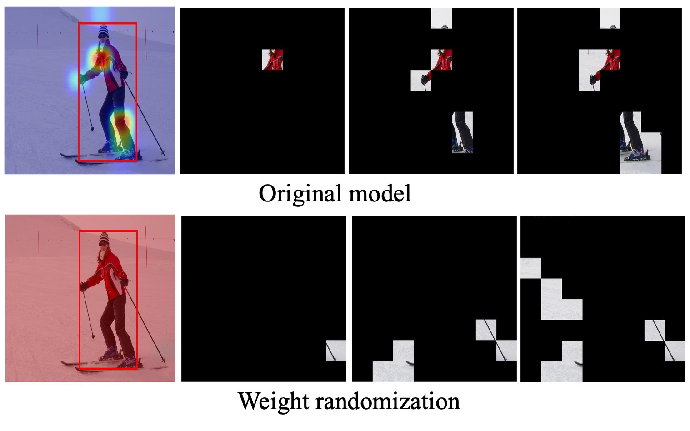}
    \caption{Comparison of heat maps generated by VX-CODE with patch insertion for the original and weight-randomized models.}
    \label{fig:sanity_checks}
\end{figure}

\begin{table}[t]
\caption{SSIM and rank correlation results under weight randomization of DETR.}
\centering
\begin{tabular}{c|cc} \toprule
\multicolumn{1}{c|}{Metric} 
& \multicolumn{1}{c}{SSIM}
& \multicolumn{1}{c}{Rank correlation} \\ \midrule

Patch insertion & 0.144 & -0.028 \\
Patch deletion & 0.312 & -0.014 \\ \bottomrule

\end{tabular}
\label{tab:sanity_checks}
\end{table}

\begin{table*}[t]
\caption{Results of pointing game (PG) and energy-based pointing game (EPG) on MS-COCO. We use bounding box (B) and instance segmentation mask (M) annotations. These results are for 1,000 detected instances from DETR with a predicted class score $>0.7$ and $\mathrm{IoU}>0.8$ with the ground truth. In VX-CODE, we set $r=1$. The best values are indicated in bold.}
\centering
\begin{tabular}{c|cccc} \toprule
Metric & PG (B) $\uparrow$ & PG (M) $\uparrow$ & EPG (B) $\uparrow$ & EPG (M) $\uparrow$ \\ \midrule

Grad-CAM     & .321 & .242 & .194 & .133\\

Grad-CAM++     & .322 & .224 & .173 & .112  \\

D-RISE     & .889 & .769 & .130 & .079 \\

SSGrad-CAM      & .618 & .490 & .314 & .224 \\

ODAM    & .602 & .455 & .315 & .215 \\

SSGrad-CAM++     & .721 & .585 & .320 & .220 \\ \midrule

VX-CODE (patch insertion)     & .951 & .905 & .520 & .351  \\ 
VX-CODE (patch deletion)     & \textbf{.965} & \textbf{.953} & \textbf{.644}& \textbf{.443}  \\ \bottomrule

\end{tabular}
\label{tab:pg}
\end{table*}

\subsection{Axiomatic properties of the \piindex}
\label{sec:axiom}
Shapley values are known to satisfy several useful axioms,
such as \textit{Null player}, \textit{Linearity}, \textit{Efficiency}, and \textit{Symmetry}.
Let $\phi_i$ be the Shapley value for player $i$ 
and $N = \{1, ..., n\}$ be the index set of players.
Each axiom is defined as follows:
\begin{itemize}
  \item \textbf{Null player}\\
       A null player is assigned a value of 0.
       A player $i$ is null in $f$ if $f(S\cup\{i\})=f(S)$.
  \item \textbf{Linearity}\\
        For any reward function $f$ and $g$ and 
        any real number $\alpha$ and $\beta$, 
        \begin{align}
            \phi_i(\alpha f + \beta g)
              &= \alpha\,\phi_i(f) + \beta\,\phi_i(g) 
        \end{align}        
  \item \textbf{Efficiency}\\
        The full value of the grand coalition is completely distributed among all players:
        \begin{align}
          \sum_{i=1}^{n} \phi_i(f) \;=\; f(N).
        \end{align}
  \item \textbf{Symmetry}\\
       If two players $i$ and $j$ are interchangeable
       (i.e., $f(S\cup\{i\}) = f(S\cup\{j\})$ for $S\subseteq N\setminus\{i,j\}$), 
        \begin{align}
          \phi_i(f)=\phi_j(f).
      \end{align}
\end{itemize}

\par
We here revisit the \piindex.
As shown in Eq.~\eqref{eq:def_ci},
in a sequential coalitional game where players are recruited into the game,
$k$-player in $\pi^*$ is assigned the \piindex as follows:
\begin{align}
    \psi(k | \pi^*) =   
         f(\pi^*(1),\ldots,\pi^*(k))-f(\pi^*(1),\ldots,\pi^*(k-1)).
\label{eq:def_ci_ins}
\end{align}
Similarly,
in a sequential coalitional game where players are removed from the game (we refer to this situation as a player elimination setting),
\piindex for this setting is defined as follows:
\begin{definition}
    Let $\pi^*\in\Pi$ be a permutation such that
    \begin{align} \label{eq:def_pi_star_del}
        \pi^* = \underset{\pi \in \Pi} {\operatorname{arg\,min}} \sum_{k=1}^{n} f(N \setminus\ \{\pi(1), \ldots, \pi(k)\}).
    \end{align}
    The $\bm{\pi}^*$\textbf{-index} of the $k$-player for the player elimination setting is defined by
    \begin{align}
            \psi^{\mathrm{es}}(k | \pi^*) & = f(N \setminus  \{\pi^*(1),\ldots,\pi^*(k-1)\}) \\
    & \quad \quad \quad \quad - f(N \setminus \{\pi^*(1),\ldots,\pi^*(k)\}).
    \label{eq:def_ci_del}
    \end{align}
\end{definition}
The $\psi^{\mathrm{es}}$ measures the contribution of each player by the decrease in reward upon their removal.
It is worth noting that the greedy patch deletion process described in App.~\ref{sec:patch_deletion}
can be interpreted as the player elimination setting in a sequential coalitional game.
As shown in \cref{eq:argmin_k2},
at each step the algorithm selects the set of patches whose removal produces the largest decrease in the reward,
which is equivalent to selecting the set with the largest full-context Shapley value.
In this sense, the patch deletion process procedure induces a specific permutation $\hat{\pi}$ that defines a sequential coalitional game in the player elimination setting.

\par
Next, 
we demonstrate that the \piindex satisfies \textit{Null player}, \textit{Linearity} and \textit{Efficiency}.
In the following, 
we implicitly assume $f(\emptyset) = 0$.
We denote the \piindex for the $k$-player with the reward function $f$ as $\psi_k(f)$.
Based on Eqs.~\eqref{eq:def_ci_ins} and \eqref{eq:def_ci_del}, 
the \piindex satisfies each axiom as follows:
\begin{itemize}
  \item \textbf{Null player}\\
          If the $k$-player is null player, 
          then we have,
          \begin{align}
              f(\pi^*(1),\ldots,\pi^*(k)) =f(\pi^*(1),\ldots,\pi^*(k-1)).
          \end{align}
          Therefore, we obtain
       \begin{align}
            \psi_k(f) & = f(\pi^*(1),\ldots,\pi^*(k))-f(\pi^*(1),\ldots,\pi^*(k-1)) \\
            & = 0.
        \end{align}
        It is straightforward to verify that the same result holds for the player elimination setting, using following fact:
          \begin{align}
              & f(N \setminus \{\pi^*(1),\ldots,\pi^*(k-1)\}) \\ 
              & \quad = f(N \setminus \{\pi^*(1),\ldots,\pi^*(k)\}).
          \end{align}
  \item \textbf{Linearity}\\
        We define a new game as a linear combination of two reward functions: $u(S) = \alpha f(S)+\beta g(S)$.
        Then, the following holds:
        \begin{align}
            & \psi_k(u) \\
            & = \psi_k(\alpha f + \beta g)  \\
          &= \alpha [f(\pi^*(1),\ldots,\pi^*(k)) - f(\pi^*(1),\ldots,\pi^*(k-1))]  \\
          & \quad + \beta [g(\pi^*(1),\ldots,\pi^*(k)) - g(\pi^*(1),\ldots,\pi^*(k-1))] \\
          & = \alpha \psi_k(f) + \beta \psi_k(g).
        \end{align}
        It is straightforward to verify that the same result holds for the player elimination setting.
  \item \textbf{Efficiency}\\
      We can show that the following holds by computing the summation of each \piindex.    
        \begin{align}
            \sum_{i=1}^{n} \psi_i(f) & = f(\pi^*(1)) - f(\emptyset) \\
            & \quad + f(\pi^*(1), \pi^*(2)) - f(\pi^*(1)) \\ 
            & \quad + \ldots  \\ 
            & \quad  +  f(\pi^*(1), \ldots, \pi^*(n)) \\
            & \quad \quad \quad \quad - f(\pi^*(1), \ldots, \pi^*(n-1)) \\ 
            & = f(\pi^*(1), ..., \pi^*(n)) - f(\emptyset) \nonumber \\ 
            & = f(N).
        \end{align}
        Similarly,
        the following holds for the player elimination setting:
        \begin{align}
            \sum_{i=1}^{n} \psi_i^{\mathrm{es}}(f) & = f(N) - f(N \setminus \pi^*(1)) \\ 
            & \quad + f(N \setminus \pi^*(1)) - f(N \setminus \{\pi^*(1), \pi^*(2)\}) \\
            & \quad + \ldots  \\ 
            & \quad + f(N \setminus \{\pi^*(1), \ldots, \pi^*(n-1)\})  \\
            & \quad\quad\quad\quad - f(N \setminus \{\pi^*(1), \ldots, \pi^*(n)\}) \\ 
            & = f(N) -  f(N \setminus \{\pi^*(1), \ldots, \pi^*(n)\}) \\ 
            & = f(N) - f(\emptyset) \\
            & = f(N).
        \end{align}
 \end{itemize}

\par
The \piindex is computed using a single permutation, $\pi^*$,
which is the optimal case that satisfies Eq.~\eqref{eq:def_pi_star} or Eq.~\eqref{eq:def_pi_star_del}.
Therefore, in the context of the $\pi^*$-index, 
the \textit{Symmetry} axiom---which is defined under averaging over all permutations---
is no longer applicable and naturally drops out.

\section{Additional experimental results}
In this section, we present additional experimental results not included in the main part.

\begin{figure*}[pt]
    \centering
    \includegraphics[scale=0.76]{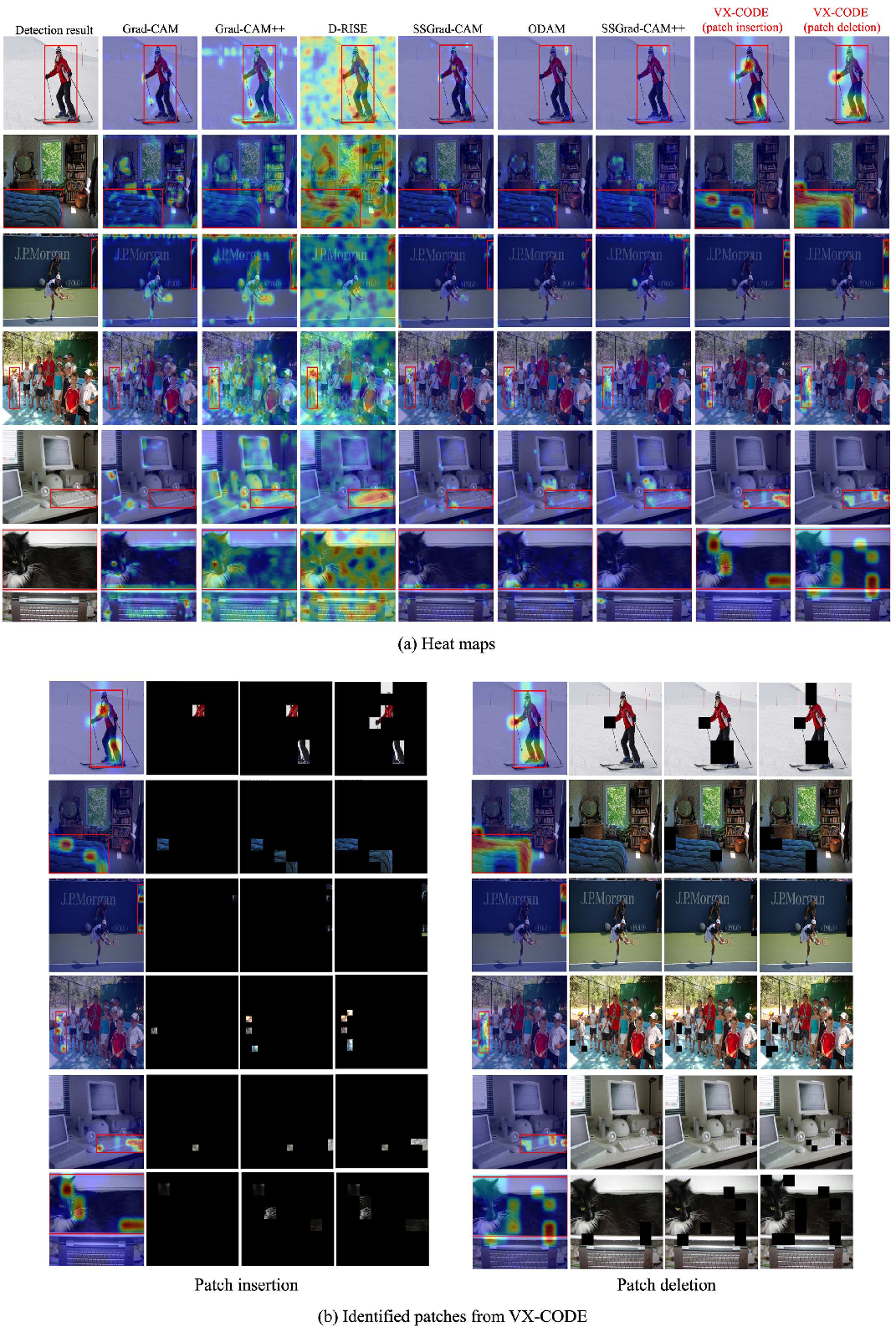}
    \caption{Additional results comparing the visual explanations generated by each method for detections from DETR. In VX-CODE, we set $r=1$. (a) Heat maps generated by each method. (b) Patches identified from VX-CODE with patch insertion (left) and patch deletion (right). Each heat map corresponds to those shown in (a).}
    \label{fig:heatmap_more1}
\end{figure*}

\begin{figure*}[pt]
    \centering
    \includegraphics[scale=0.78]{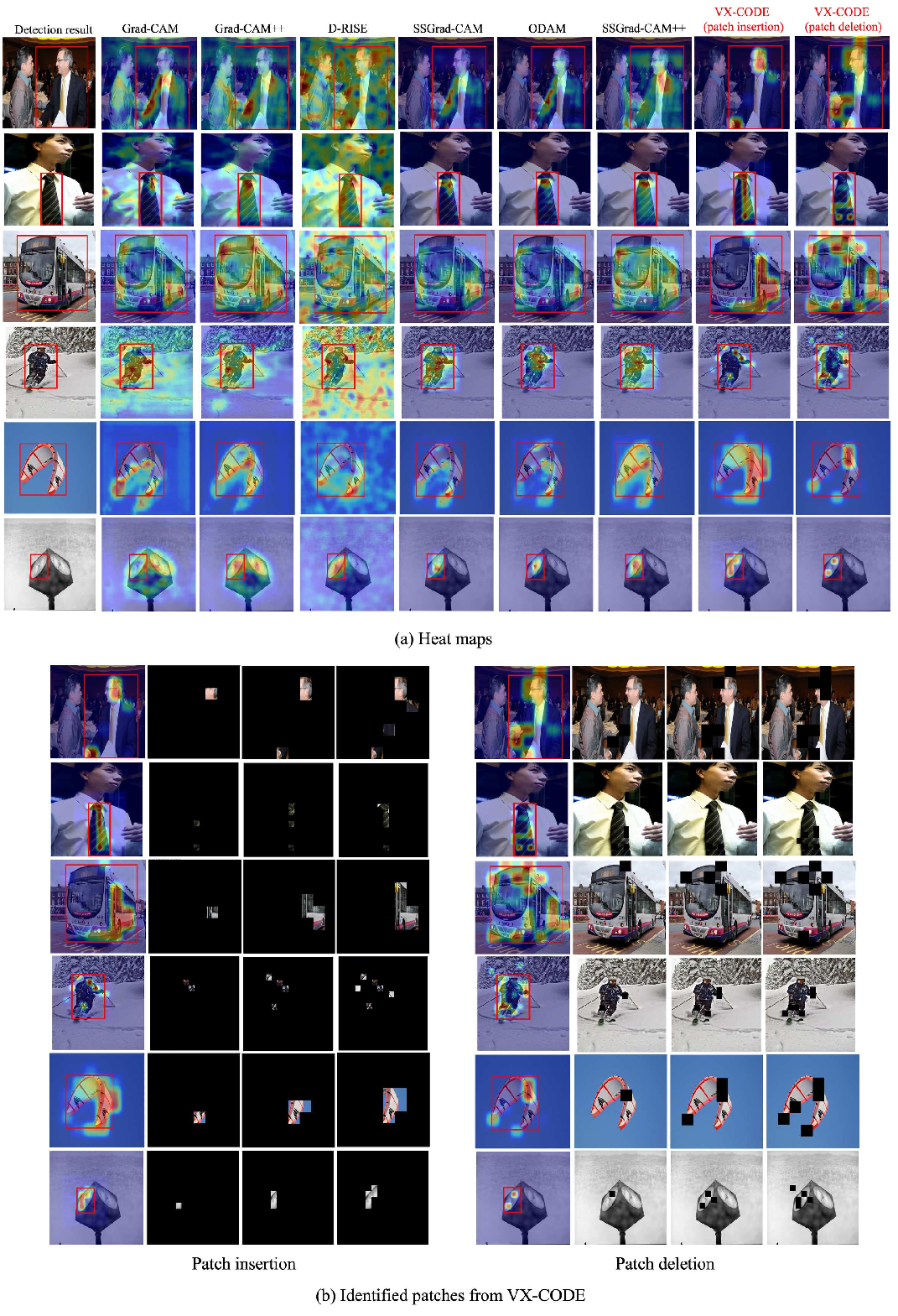}
    \caption{Additional results comparing the visual explanations generated by each method for detections from Faster R-CNN. In VX-CODE, we set $r=1$. (a) Heat maps generated by each method. (b) Patches identified from VX-CODE with patch insertion (left) and patch deletion (right). Each heat map corresponds to those shown in (a).}
    \label{fig:heatmap_more2}
\end{figure*}

\begin{figure*}[t]
    \centering
    \includegraphics[scale=0.4]{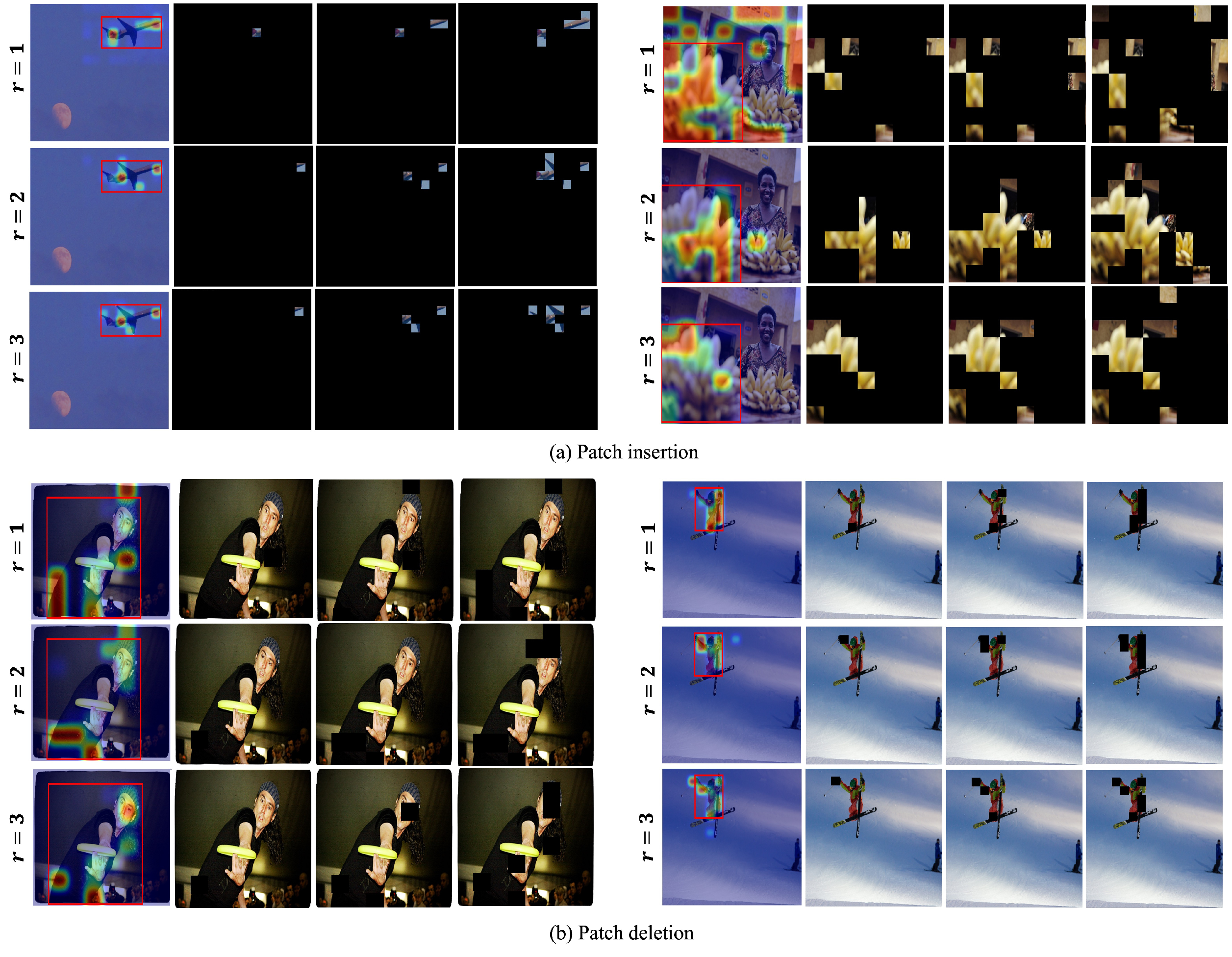}
    \caption{Additional results comparing the identified patches and generated heat maps by VX-CODE when changing $r$. (a) Results for patch insertion. (b) Results for patch deletion.}
    \label{fig:change_a_more}
\end{figure*}

\begin{figure}[t]
    \centering
    \includegraphics[scale=0.5]{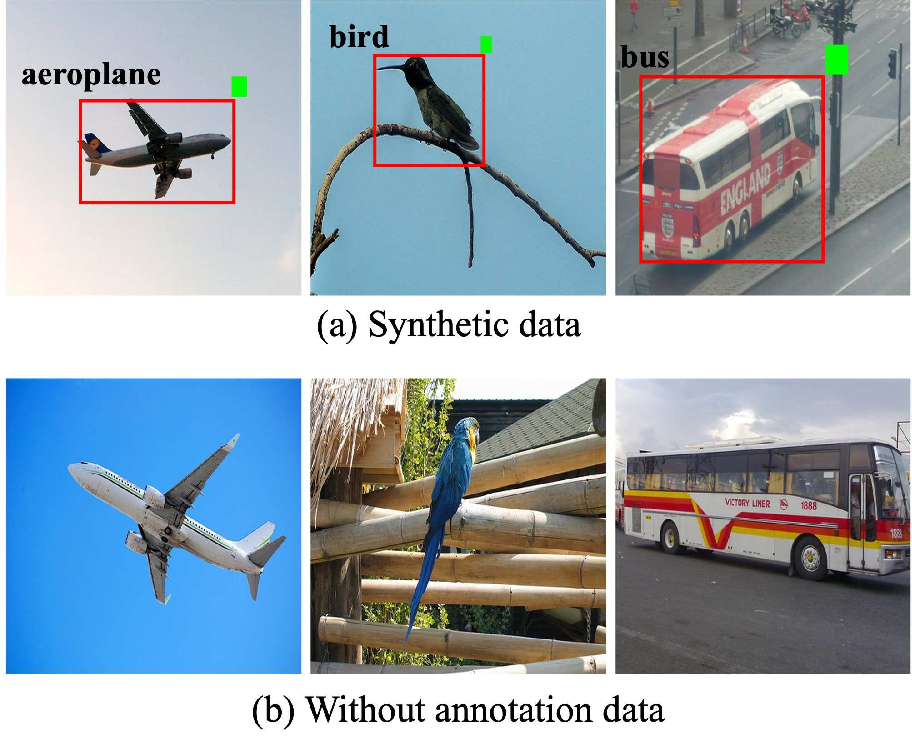}
    \caption{(a) Example of synthetic data: A square marker is placed in the upper-right corner outside the bounding box. These data include annotations for the class and bounding box. (b) Example of data without annotations: These data lack annotations and are treated as background during training.}
    \label{fig:bias_data}
\end{figure}

\subsection{Comparison of filling methods}
\label{sec:zero_blur}
As described in Sec.~\ref{sec:experiments}, 
we fill the masked regions with zeros in the proposed method. 
Here, we compare two filling methods,
zero filling and blurred pixel values,
using the insertion and deletion metrics.
Figure~\ref{fig:zero_blur} shows the results. 
We observe that both methods produce nearly identical results. 
These findings suggest that the choice of filling method has no significant impact on the proposed method.

\subsection{Comparison of Monte-Carlo based Shapley values}
\label{sec:compare_shap}
We compare the proposed method with Shapley values estimated via Monte Carlo sampling.
Figure~\ref{fig:compare_shap} shows the results of the insertion and deletion metrics, computed for 100 instances detected by DETR on MS-COCO. 
The number of Monte Carlo samples is set to 200, 
following prior studies,
which already incurs a higher computational cost than our method with $r=1$. 
As shown in Fig.~\ref{fig:compare_shap}, 
the proposed method with $r=1$ already outperforms the Shapley values estimated by Monte Carlo sampling.

\subsection{Sanity checks}
\label{sec:sanity_checks}
We conduct sanity checks to evaluate the sensitivity of the proposed method to the object detectors.
Sanity checks assess whether each explanation method provides meaningful explanations~\cite{Adebayo2018, sanity2}.
Specifically, they examine how the outputs of the methods change when model parameters and labels are randomized.
If a method lacks sensitivity to such randomizations and produces results similar to those of edge detectors,
it suggests that the method does not rely on the training data or the model~\cite{Adebayo2018}.
Consequently, such methods are inadequate for tasks such as identifying outliers in the data, as they are independent of both the model and the data.

\par
We perform the checks using model weight randomization.
Figure~\ref{fig:sanity_checks} shows an example generated by VX-CODE with patch insertion for DETR.
The proposed method identifies uninformative patches under randomized weights.
Table~\ref{tab:sanity_checks} presents the SSIM and rank correlation results computed for 100 detected instances using DETR on MS-COCO.
These metrics evaluate the similarity between the explanations for the original and randomized models (with values closer to 1.0 indicating higher similarity).
We observe significant drops in these metrics.
These results indicate that the proposed method provides meaningful explanations that are sensitive to the model’s parameters.

\begin{table*}[t]
\caption{Results of AUC for insertion (Ins), deletion (Del), and overall (OA) metrics for the classes {\it aeroplane}, {\it bird}, {\it bottle}, and {\it bus}. In VX-CODE, we set $r=1$.
Each result is computed over 50 detection instances. The best values are indicated in bold.}
\centering
\begin{tabular}{c|ccc|ccc|ccc|ccc} \toprule
\multicolumn{1}{c}{Class} 
& \multicolumn{3}{c}{{\it aeroplane}} 
& \multicolumn{3}{c}{{\it bird}} 
& \multicolumn{3}{c}{{\it bottle}} 
& \multicolumn{3}{c}{{\it bus}} \\ \cmidrule{2-13}
\multicolumn{1}{c}{Metric} & Ins $\uparrow$ & Del $\downarrow$ & \multicolumn{1}{c}{OA} $\uparrow$ 
& Ins $\uparrow$ & Del $\downarrow$ & \multicolumn{1}{c}{OA} $\uparrow$
& Ins $\uparrow$ & Del $\downarrow$ & \multicolumn{1}{c}{OA} $\uparrow$ 
& Ins $\uparrow$ & Del $\downarrow$ & \multicolumn{1}{c}{OA} $\uparrow$ \\ \midrule

Grad-CAM 
& .517 & .026 & .491
& .601 & .091 & .510
& .740 & .017 & .723 
& .534 & .148 & .386 \\

Grad-CAM++
& .658 & .016 & .642
& .718 & \textbf{.022} & .696
& .853 & .011 & .842
& .692 & .028 & .664 \\

D-RISE
& .680 & \textbf{.014} & .666
& .672 & .025 & .647
& .843 & \textbf{.009} & .834
& .581 & \textbf{.023} & .558 \\

SSGrad-CAM
& .649 & .053 & .596
& .641 & .078 & .563
& .803 & .017 & .786
& .629 & .090 & .539 \\

ODAM
& .400 & .089 & .311
& .470 & .074 & .396
& .516 & .048 & .468
& .502 & .091 & .411 \\

SSGrad-CAM++ 
& .796 & .044 & .752
& \textbf{.822} & .061 & .761
& .902 & .014 & .888
& .768 & .050 & .718 \\ \midrule

VX-CODE 
& \textbf{.891} & .034 & \textbf{.857}
& .807 & .035 & \textbf{.772}
& \textbf{.904} & .014 & \textbf{.890}
& \textbf{.873} & .059 & \textbf{.814} \\ \bottomrule

\end{tabular}
\label{tab:del_ins_bias_more1}
\end{table*}

\begin{table*}[t]
\caption{Results of AUC for insertion (Ins), deletion (Del), and overall (OA) metrics for the classes {\it dog}, {\it motorbike}, and {\it sofa}. In VX-CODE, we set $r=1$.
Each result is computed over 50 detection instances. The best values are indicated in bold.}
\centering
\begin{tabular}{c|ccc|ccc|ccc} \toprule
\multicolumn{1}{c}{Class} 
& \multicolumn{3}{c}{{\it dog}} 
& \multicolumn{3}{c}{{\it motorbike}} 
& \multicolumn{3}{c}{{\it sofa}} \\ \cmidrule{2-10}
\multicolumn{1}{c}{Metric} & Ins $\uparrow$ & Del $\downarrow$ & \multicolumn{1}{c}{OA} $\uparrow$ 
& Ins $\uparrow$ & Del $\downarrow$ & \multicolumn{1}{c}{OA} $\uparrow$ 
& Ins $\uparrow$ & Del $\downarrow$ & \multicolumn{1}{c}{OA} $\uparrow$ \\ \midrule

Grad-CAM
& .605 & .046 & .559
& .498 & .199 & .299
& .465 & .071 & .394 \\

Grad-CAM++
& .697 & .019 & .678
& .750 & .023 & .727
& .620 & .011 & .609 \\

D-RISE
& .690 & \textbf{.014} & .676
& .679 & \textbf{.013} & .666
& .655 & \textbf{.008} & .647 \\

SSGrad-CAM
& .591 & .076 & .515
& .682 & .118 & .564
& .524 & .054 & .470 \\

ODAM
& .510 & .091 & .419
& .535 & .142 & .393
& .460 & .125 & .335 \\

SSGrad-CAM++
& .772 & .050 & .722
& .803 & .056 & .747
& .747 & .041 & .706 \\ \midrule

VX-CODE 
& \textbf{.840} & .040 & \textbf{.800}
& \textbf{.899} & .039 & \textbf{.860}
& \textbf{.852} & .034 & \textbf{.818} \\ \bottomrule

\end{tabular}
\label{tab:del_ins_bias_more2}
\end{table*}

\subsection{Pointing game}
\label{sec:pg}
To evaluate the localization ability of the generated heat map, 
we conduct evaluations using pointing game \cite{pointing_game} and energy-based pointing game \cite{score_cam}.
The pointing game measures the proportion of heat maps in which the maximum value falls within the ground truth, such as a bounding box or instance mask. 
Meanwhile, the energy-based pointing game assesses the proportion of the heat map energy that lies within the ground truth.
Note that these metrics only evaluate the localization ability of the generated heat map with respect to the ground truth instance rather than the faithfulness of the heat map. 
For example, as discussed in the main paper, 
object detectors may consider regions outside the bounding box when detecting objects, rendering these metrics potentially less effective in such cases.

\par
Table~\ref{tab:pg} summarizes the results for those metrics.
We compute each metric for the 1,000 instances detected by DETR on MS-COCO,
with a predicted class score $>0.7$ and $\mathrm{IoU} > 0.8$ with the ground truth.
The results for VX-CODE are computed using the heat maps generated from patch insertion and patch deletion with $r=1$.
As shown in Tab.~\ref{tab:pg},
both VX-CODE with patch insertion and patch deletion demonstrate high localization ability for the detected instances.
Notably, VX-CODE with patch deletion outperforms patch insertion across these metrics.
As described in Sec.~\ref{sec:compare_heatmap},
patch deletion selects patches that reduce the class score, as defined in Eq.~\eqref{eq:reward},
This strategy focuses on eliminating critical instance features, enhancing its ability to localize objects effectively. 
These results support the advantages of patch deletion for object localization.

\subsection{Additional results for visualizations}
\label{sec:more_heatmap}
In Sec.~\ref{sec:compare_heatmap},
we show the explanations generated by each method.
Figure~\ref{fig:heatmap_more1} and Fig.~\ref{fig:heatmap_more2} provide the additional examples of explanations generated by each method for detections from DETR and Faster R-CNN, respectively.
Each figure includes the heat maps generated by each method and the patches identified by VX-CODE in the early steps.
As described in Sec.~\ref{sec:compare_heatmap},
the proposed method identifies the important patches for detections by considering the collective contributions of multiple patches.

\subsection{Additional results for patch combination}
\label{sec:more_patch_combination}
In Sec.~\ref{sec:compare_heatmap}, 
we demonstrate the effect of increasing $r$.
Here, we show the additional results to further support this analysis.
Figure~\ref{fig:change_a_more} illustrates the effect of $r$ on patch identification. 
As shown in Fig.~\ref{fig:change_a_more}, 
the proposed method identifies more features, reflecting the effect of considering the collective contributions of a larger set of patches.
For example, as shown on the left side of Fig.~\ref{fig:change_a_more} (a) (detection for the aeroplane), 
the identified features for $r=1$ include the front and rear parts of the aeroplane,
while the wings are additionally identified for $r=2$ and $r=3$, 
indicating that these features collectively contribute to the detection.

\par
Increasing $r$ also leads to more accurate identification of important patches.
As shown on the right side of Fig.~\ref{fig:change_a_more} (a) (detection of the banana), 
the case with $r=1$ fails to identify the critical patches in the initial steps, 
whereas $r=2$ and $r=3$ successfully identify these patches at earlier steps.
This improvement is particularly effective for patch insertion.
Patch insertion gradually adds patches to an empty image. 
Therefore, at earlier steps, 
the reward function value changes only slightly due to the limited pixel information, 
making it difficult to identify important patches.
Increasing $r$ mitigates this problem by considering Shapley values and interactions among more patches, 
leading to more accurate identification of important patches.

\begin{figure*}[t]
    \centering
    \includegraphics[scale=0.8]{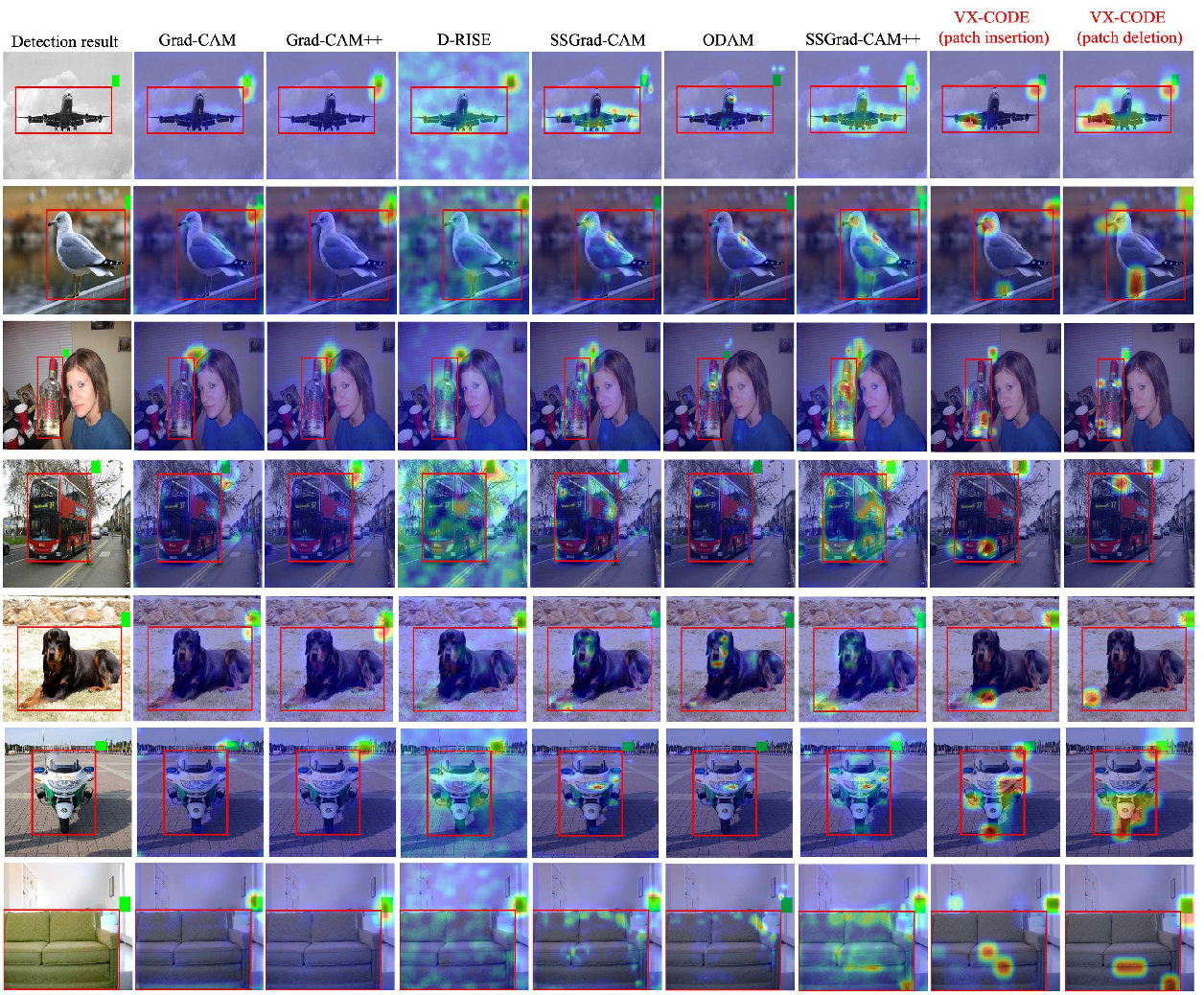}
    \caption{Additional results comparing the visualizations generated by each method for the biased model. In VX-CODE, we set $r=1$. The results correspond to the classes {\it aeroplane}, {\it bird}, {\it bottle}, {\it bus}, {\it dog}, {\it motorbike}, and {\it sofa}, listed from the top row downward.}
    \label{fig:bias_more1}
\end{figure*}

\begin{figure*}[t]
    \centering
    \includegraphics[scale=0.8]{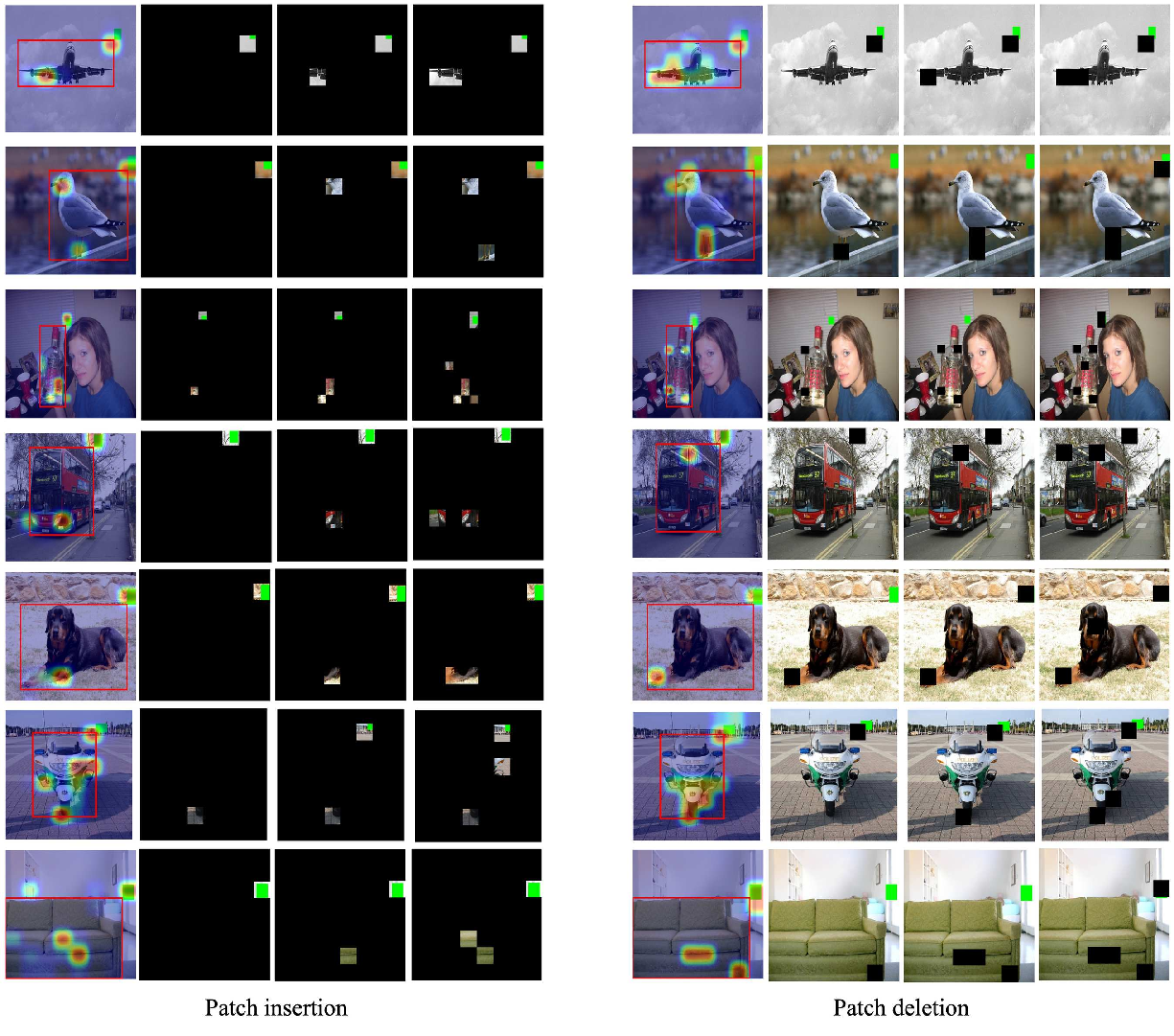}
    \caption{Patches identified from VX-CODE with patch insertion (left) and patch deletion (right) corresponding to the heat maps shown in Fig.~\ref{fig:bias_more1}.}
    \label{fig:bias_more2}
\end{figure*}

\subsection{Additional results for biased model}
\label{sec:more_bias_model}
In Sec.~\ref{sec:problem_setup},
we demonstrate the effectiveness of the proposed method through experiments using a biased model. Here, we provide details about the experimental setup, 
including the synthetic dataset and the trained model, along with additional results.

\par
Figure~\ref{fig:bias_data} illustrates an example of the constructed synthetic dataset.
As shown in Fig.~\ref{fig:bias_data} (a),
a square marker is placed outside the upper-right corner of the bounding box for seven classes: {\it aeroplane}, {\it bird}, {\it bottle}, {\it bus}, {\it dog}, {\it motorbike}, and {\it sofa}.
These data include annotations for the corresponding class and its bounding box.
In our constructed dataset, these synthetic data account for approximately 70\%.
In the remaining 30\%, 
data corresponding to each class but without annotations are included and are treated as background during training.
DETR trained on such data considers both the features of the marker outside the bounding box and those of the instance itself to detect objects and distinguish classes appropriately.
In practice, 
this model achieves a mean average precision (mAP@50) with an IoU threshold $>0.5$ of 83.41 for the seven classes on the synthetic test dataset.
On the other hand, the mAP@50 drops to 34.73 on the original test dataset, where the square marker is not placed.
These results indicate that DETR trained on the constructed synthetic dataset is heavily biased toward the square marker outside the bounding box.

\par
We present results for the deletion, insertion, and overall metrics for each class in Tab.~\ref{tab:del_ins_bias_more1} and Tab.~\ref{tab:del_ins_bias_more2}.
Table~\ref{tab:del_ins_bias_more1} shows the results for the classes {\it aeroplane}, {\it bird}, {\it bottle}, and {\it bus},
and Tab~\ref{tab:del_ins_bias_more2} presents the results for the classes {\it dog}, {\it motorbike}, and {\it sofa}. 
Each result is computed over 50 detected instances.
The proposed method outperforms existing methods, 
particularly in the insertion metric, 
where both instance and marker features must be identified for correct detection.
As shown in Fig.~\ref{fig:heatmap_bias_auc}, 
the proposed method effectively identified these two regions.
These results show that VX-CODE provides explanations that consider both the instance and marker features in the detection.
Interestingly, 
D-RISE and Grad-CAM++ marginally outperform VX-CODE in the deletion metric.
In this metric, 
the similarity score significantly decreases when either the instance or the marker is identified.
As shown in Fig.~\ref{fig:heatmap_bias_auc}, 
D-RISE and Grad-CAM++ strongly highlight only the marker, 
contributing to their better performance. 
However, these methods fail to capture the instance features, 
leading to poor performance in the insertion metric.

\par
Figure~\ref{fig:bias_more1} shows additional examples of visualizations generated by each method for the biased model,
and Fig.~\ref{fig:bias_more2} shows the patches identified from VX-CODE. 
As discussed in Sec~\ref{sec:problem_setup}, 
the proposed method efficiently identifies the features of both the instance and the marker, 
whereas existing methods tend to highlight only the marker (e.g., the result of Grad-CAM++ and D-RISE for the detection of the dog) or emphasize all features (e.g., the result of SSGrad-CAM++ for the detection of the bottle).

\begin{figure*}[ht]
    \centering
    \includegraphics[scale=0.6]{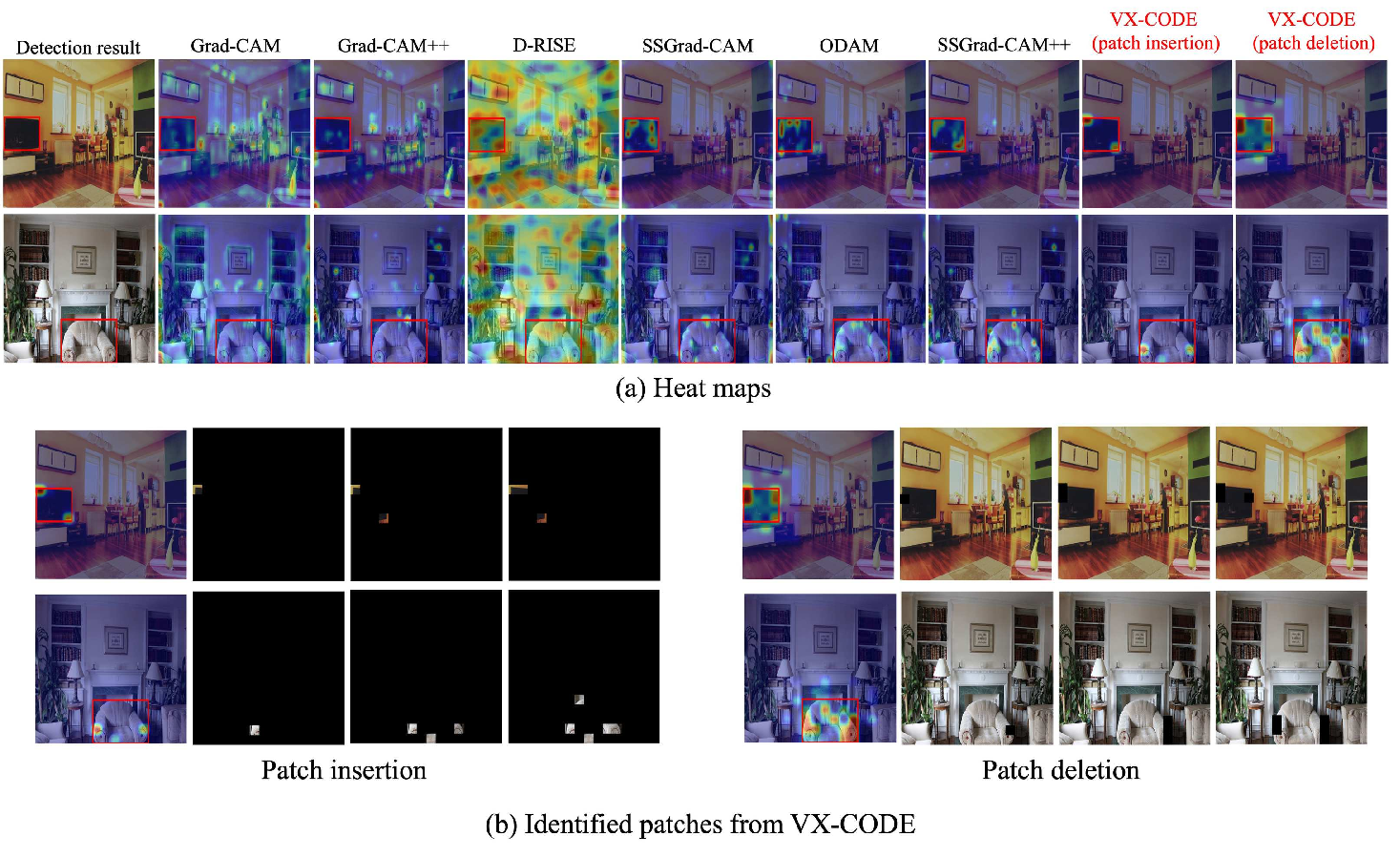}
    \caption{Comparison results of the visual explanations for bounding box predictions generated by each method. In VX-CODE, we set $r=1$. (a) Heat maps generated by each method. (b) Patches identified from VX-CODE with patch insertion (left) and patch deletion (right). Each heat map corresponds to those shown in (a).}
    \label{fig:box_more}
\end{figure*}

\begin{figure*}[ht]
    \centering
    \includegraphics[scale=0.4]{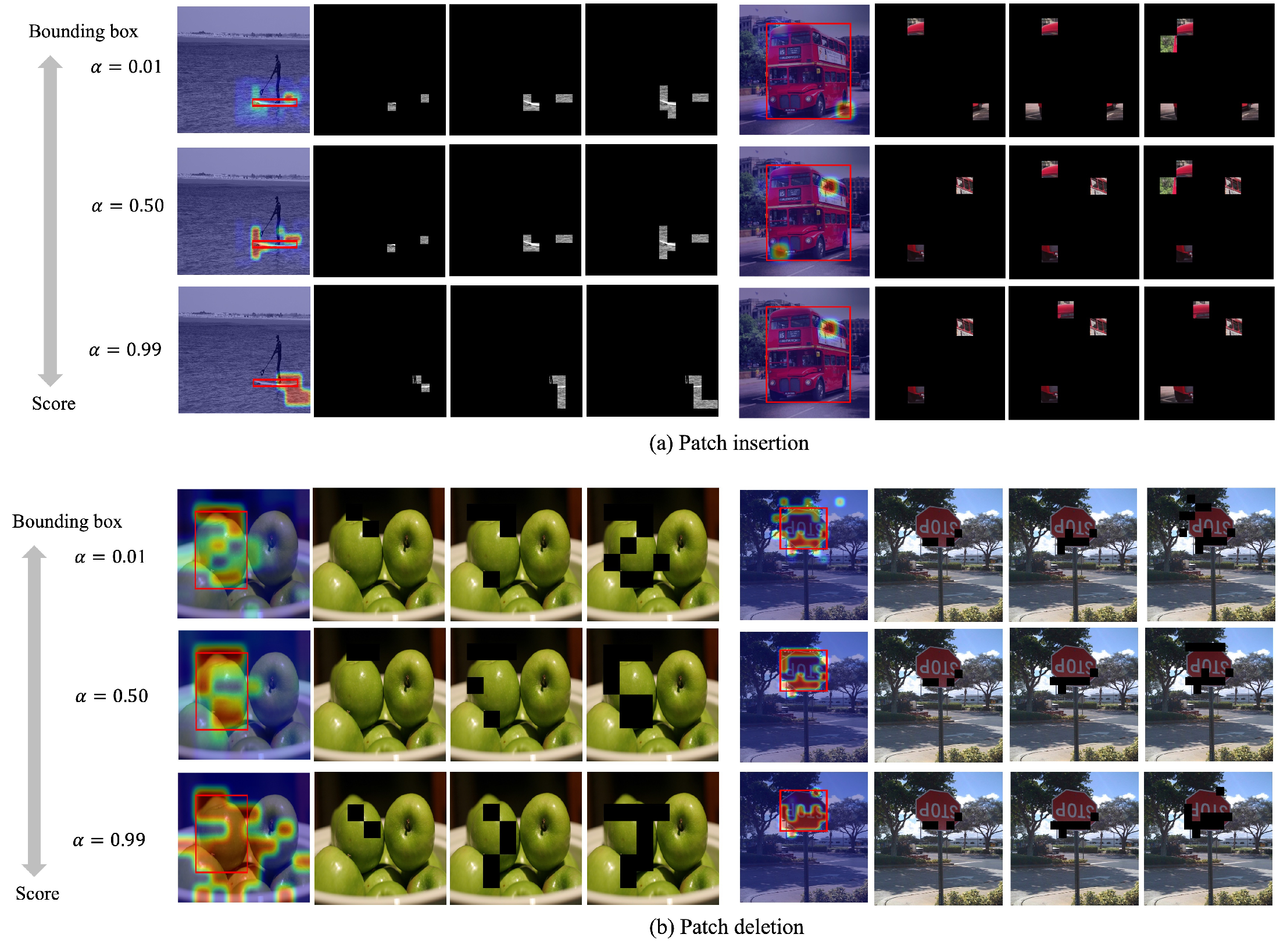}
    \caption{The differences in regions identified by VX-CODE when varying $\alpha$ in Eq.~\eqref{eq:reward_change_a}, with (a) patch insertion and (b) patch deletion.}
    \label{fig:change_reward}
\end{figure*}

\begin{figure*}[ht]
    \centering
    \includegraphics[scale=0.6]{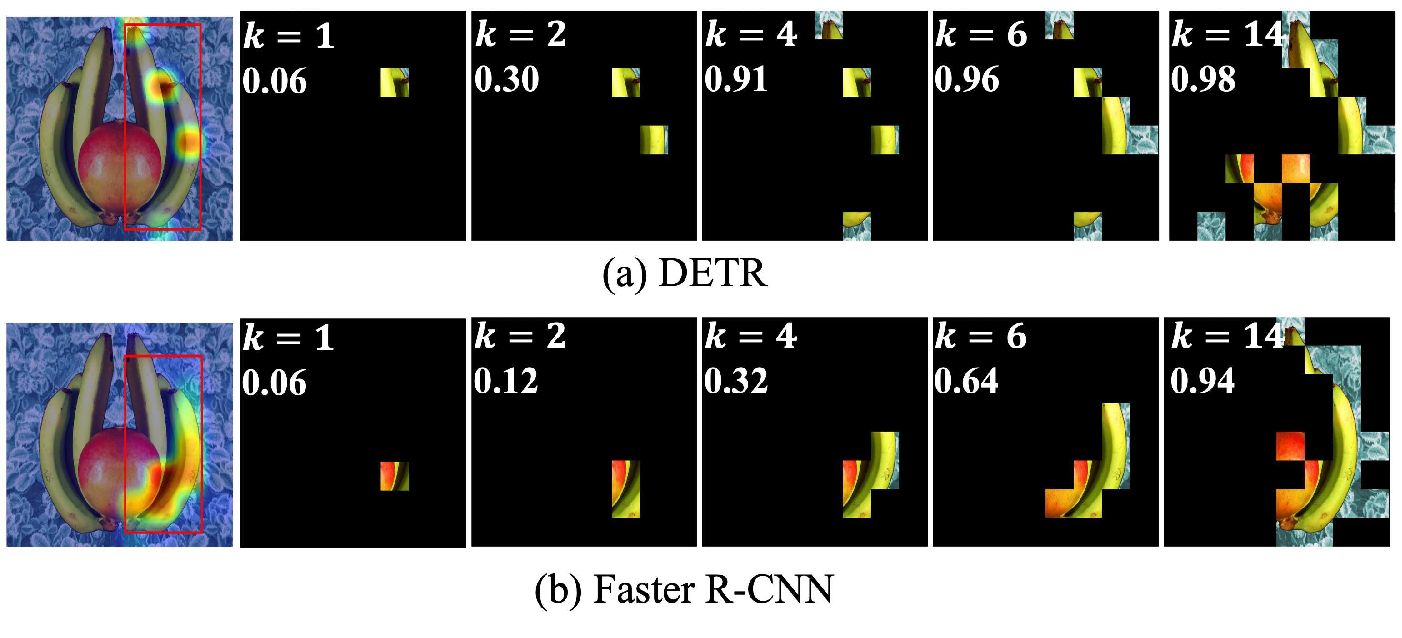}
    \caption{The differences in identified regions by VX-CODE with patch insertion between (a) DETR and (b) Faster R-CNN. Each figure shows the number of identified patches $k$ and the corresponding reward value.}
    \label{fig:detr_faster}
\end{figure*}

\begin{table}[t]
\caption{Results of AUC for insertion (Ins), deletion (Del), and overall (OA) metrics. In VX-CODE, we set $r=1$.
Each result is computed over 1,000 detection instances with a predicted class score above $0.7$ from Faster R-CNN on MS-COCO. The best values are indicated in bold.}
\centering
\begin{tabular}{c|ccc} \toprule
Metric & Ins $\uparrow$ & Del $\downarrow$ & OA $\uparrow$ \\ \midrule

Grad-CAM  & .627 & .234 & .393 \\

Grad-CAM++  & .837 & .115 & .722 \\

D-RISE  & .864 & .089 & .775 \\

SSGrad-CAM  & .819 & .078 & .741\\

ODAM & .866 & .077 & .789 \\

SSGrad-CAM++  & .898 & .060 & .838\\ \midrule

VX-CODE & \textbf{.922} & \textbf{.025} & \textbf{.897} \\ \bottomrule

\end{tabular}
\label{tab:del_ins_score}
\end{table}

\begin{table}[t]
\caption{Results of AUC for insertion (Ins), deletion (Del), and overall (OA) metrics. In VX-CODE, we set $r=1$.
Each result is computed over 1,000 detection instances with a predicted class score above $0.7$ from DETR on MS-COCO. The best values are indicated in bold.}
\centering
\begin{tabular}{c|ccc} \toprule
Metric & Ins $\uparrow$ & Del $\downarrow$ & OA $\uparrow$ \\ \midrule

Grad-CAM     & .799 & .474 & .325 \\

Grad-CAM++     & .729 & .521 & .208  \\

D-RISE     & .881 & .397 & .484 \\

SSGrad-CAM      & .884 & .400 & .484\\

ODAM     & .885 & .383 & .502  \\

SSGrad-CAM++     & .864 & .401 & .463\\ \midrule

VX-CODE     & \textbf{.926} & \textbf{.238} & \textbf{.688}  \\ \bottomrule

\end{tabular}
\label{tab:del_ins_box}
\end{table}

\subsection{Quantitative evaluation for class explanations}
\label{sec:del_ins_class_score}
We focus on explanations specifically for the predicted class.
For this purpose, the reward function in VX-CODE is
defined as the probability $p_y \in [0,1]$ of the predicted class $y$ for the original detected instance.

\par
Table~\ref{tab:del_ins_score} shows the insertion, deletion, and overall metrics,
based on 1,000 instances detected by Faster R-CNN on MS-COCO.
The proposed method outperforms the other methods,
demonstrating its ability to accurately identify the important regions for the predicted class.

\begin{figure*}[ht]
    \centering
    \includegraphics[scale=0.7]{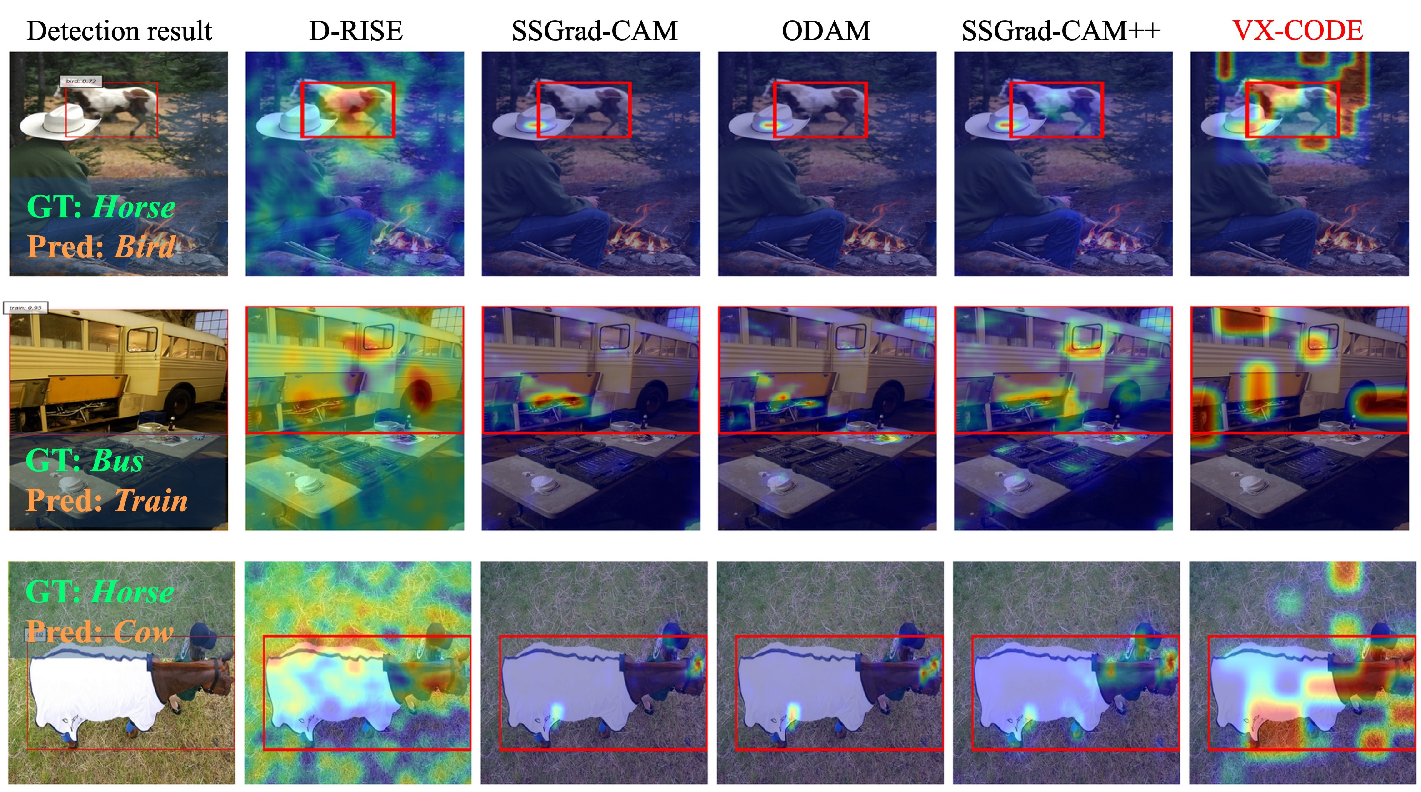}
    \caption{Visualization results for misclassification generated by D-RISE, SSGrad-CAM, ODAM, SSGrad-CAM++, and VX-CODE with patch insertion.
    In the first row, the horse is misclassified as a bird; 
    in the second row, the bus is misclassified as a train; 
    and in the third row, the horse is misclassified as a dog.}
    \label{fig:failure_more_class}
\end{figure*}

\begin{figure*}[ht]
    \centering
    \includegraphics[scale=0.7]{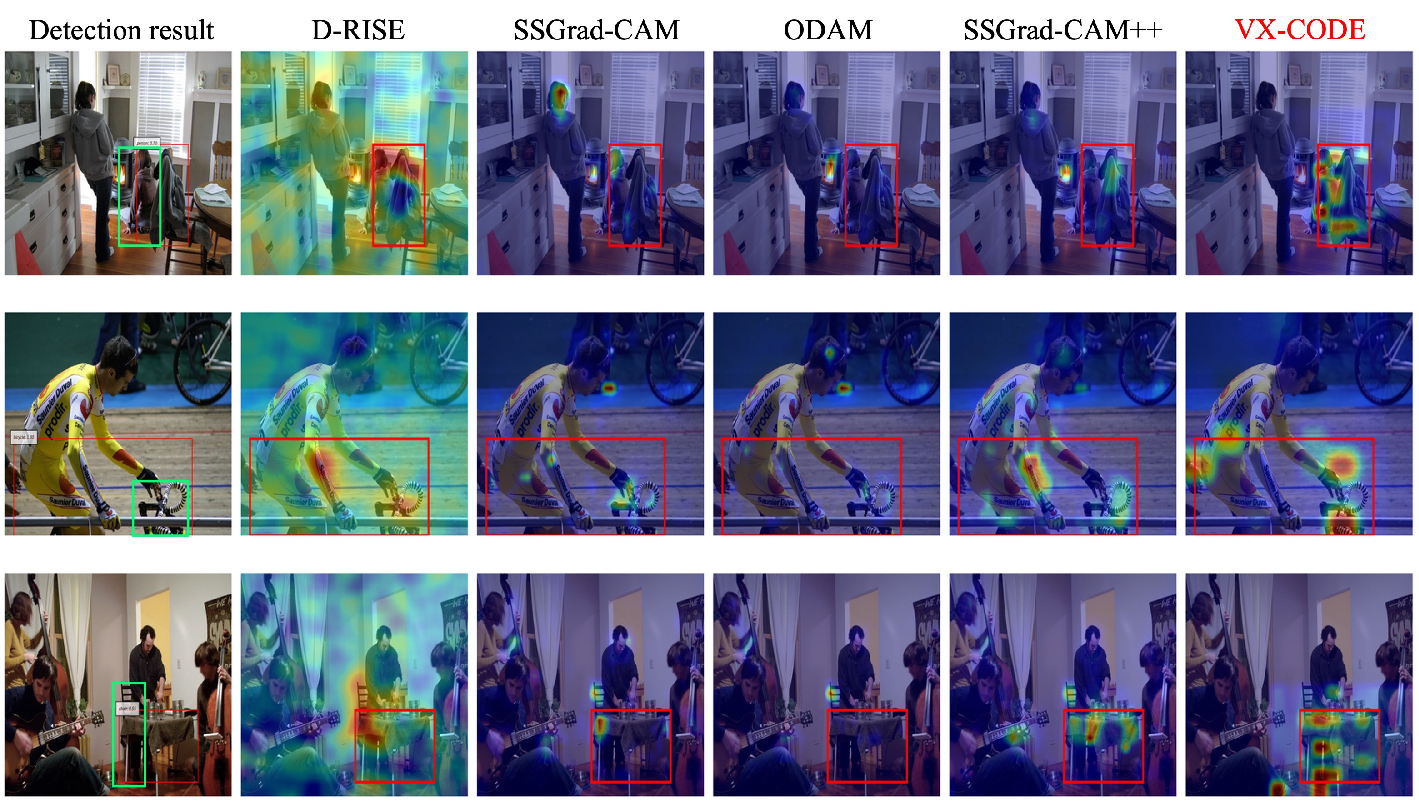}
    \caption{Visualization results for mislocalization generated by D-RISE, SSGrad-CAM, ODAM, SSGrad-CAM++, and VX-CODE with patch insertion.
    In each detection result, the green box represents the ground truth bounding box, 
    while the red box represents the predicted bounding box.
    The predicted labels, from top to bottom, are {\it person}, {\it bicycle}, and {\it chair}.}
    \label{fig:failure_more_box}
\end{figure*}

\subsection{Analysis for bounding box explanations}
\label{sec:box_explanation}
We evaluate explanations for bounding box generation.
In this experiment, to generate explanations solely for the bounding boxes, 
we set the following reward function for VX-CODE and D-RISE as a variant of Eq.~\eqref{eq:reward}.
\begin{align}\label{eq:reward_box}
    f(\mathcal{D}(x); L^{x}) = \underset{L \in \mathcal{D}(x)} {\operatorname{max}} \mathrm{IoU}(L^{x},L),
\end{align}
where $\mathcal{D}$ is the object detector, $L^{x}$ is the predicted bounding box for the original image,
and $\mathrm{IoU}(L^{x},L)$ denotes the IoU between the proposal bounding box $L$ and $L^{x}$.
For methods utilizing gradients (Grad-CAM, Grad-CAM++, SSGrad-CAM, ODAM, and SSGrad-CAM++), 
we first generate the heat maps for each bounding box coordinate in $\{x_1, y_1, x_2, y_2\}$ following~\cite{odam}.
Then, we obtain the heat map $H_{\mathrm{box}}$ by normalizing these heat maps and summing them together, as follows:
\begin{align}
    H_{\mathrm{box}} = \sum_{i\in \{x_1, y_1, x_2, y_2\}} \frac{H_i}{\max(H_i)},
    \label{eq:heatmap_box}
\end{align}
where $H_i \in \mathbb{R}^{H \times W}$ is the heat map for each bounding box coordinate on a feature map of size $H \times W$, 
and $\max(\, \cdot \,)$ calculates the maximum value of a matrix.

\par
Table~\ref{tab:del_ins_box} shows the results of insertion, deletion, and overall metrics, 
based on 1,000 instances from the results of DETR on MS-COCO.
Here, we use Eq.~\eqref{eq:reward_box} as the similarity score calculated at each step.
The proposed method outperforms the others, demonstrating its ability to accurately identify the important regions for the bounding box generation.
Figure~\ref{fig:box_more} compares the visualization results.
Generally, the size of a bounding box is determined by several points (at least two). 
Therefore, considering their collective contributions is important.
As shown in Fig.~\ref{fig:box_more}, 
the proposed method accurately identifies important regions by considering such collective contributions. 
In contrast, D-RISE, which uses the same reward function, 
and methods utilizing gradients generate noisy results or also highlight irrelevant regions.
These results support its superiority in the quantitative results shown in Tab.~\ref{tab:del_ins_box}.

\subsection{Multi-perspective explanations for detection results}
\label{sec:change_reward}
The proposed method identifies important patches based on a reward function.
One advantage of the proposed method is that it allows for the generation of multi-perspective explanations for a detection result through the adjustment of this reward function.
To this end,
we weight each term in the reward function of Eq.~\eqref{eq:reward} with $\alpha \in [0,1]$ as follows:
\begin{align}\label{eq:reward_change_a}
    & f(\mathcal{D}(x); (L^{x}, P^{x})) = \\
    & \underset{(L, P) \in \mathcal{D}(x)} {\operatorname{max}} \left\{\mathrm{IoU}(L^{x}, L)\right\}^{1-\alpha} \cdot \left\{\frac{P^{x} \cdot P}{\|P^{x}\| \|P\|}\right\}^{\alpha}.
\end{align}
As shown in this equation, 
if $\alpha$ is close to $0$, 
it identifies patches that focus more on the prediction of bounding boxes.
If $\alpha$ is close to $1$,
it identifies patches that focus more on the prediction of class scores.
If $\alpha=0.5$,
it identifies patches that balance their focus between the prediction of class scores and bounding boxes, 
which is essentially the same as using Eq.~\eqref{eq:reward}.

\par
Figure~\ref{fig:change_reward} shows the identified patches and the generated heat maps from VX-CODE with (a) patch insertion and (b) patch deletion in the cases where $\alpha$ is $0.01$, $0.50$, and $0.99$
These examples illustrate how the focus shifts between bounding box prediction and class prediction depending on the weight in the reward function.
For example, as shown in the left example of (a) patch insertion, 
for the detection of the surfboard with $\alpha=0.01$, 
the proposed method preferentially highlights area near the front and rear of the surfboard, 
indicating that these features significantly contribute to the bounding box prediction.
In the case with $\alpha=0.99$, 
the proposed method preferentially highlights the rear of the surfboard and the sea outside the bounding box, 
indicating that these features significantly contribute to the class ({\it surfboard}) prediction.
In the case with $\alpha=0.50$,
the proposed method highlights these features (the front and rear of the surfboard and the sea outside the bounding box) in a balanced manner.
A similar trend can be seen in the examples of (b) patch deletion.
For example, as shown in the left example, 
for the detection of the apple with $\alpha=0.01$, 
the proposed method preferentially highlights regions near the bounding box outline.
In the case with $\alpha=0.99$, 
the proposed method not only highlights the detected apple but also other apples. 
This identification is natural because the selection focuses on the class prediction of {\it apple}, 
and these features influence it in other proposals generated by the object detector.
In the case with $\alpha=0.50$, 
the proposed method only highlights features of the apple within the detected instance.
As shown in these examples, 
the proposed method can adjust its focus between bounding box generation and class prediction by adjusting the weight in the reward function, 
enabling the generation of multi-perspective explanations.

\subsection{Analysis of object detectors using VX-CODE}
\label{sec:analysis_detector}
We present an example of analyzing object detectors using explanations generated by VX-CODE.
Here, we analyze the differences in behavior between transformer-based and CNN-based architectures.
Several studies have explored the differences in characteristics between transformers and CNNs, 
suggesting that transformers utilize more global information and exhibit more compositionality compared to CNNs~\cite{vit_cnn1, vit_cnn2, decision_making}.
In this experiment, to analyze such differences in object detectors,
we use VX-CODE with patch insertion to identify important regions for the same detection results detected by DETR and Faster R-CNN.

\par
Figure~\ref{fig:detr_faster} shows the identified patches and generated heat maps for detections of a banana produced by DETR and Faster R-CNN.
For DETR, patches in multiple parts are identified in the early steps, 
and the reward value defined in Eq.~\eqref{eq:reward} reaches $0.91$ after four patches are identified.
In contrast, for Faster R-CNN, patches in specific regions are preferentially identified, 
and the reward value reaches $0.94$ after 14 patches are identified.
These results indicate that DETR, a transformer-based architecture, recognizes instances in a more compositional manner and with less information compared to the CNN-based architecture Faster R-CNN.

\par
As demonstrated in this experiment, 
the explanations obtained from VX-CODE facilitate the analysis of object detectors. 
Further investigations into such analyses using the proposed method are left for future work.

\subsection{Additional results for failure cases}
\label{sec:eval_failer_case}
In Secs.~\ref{sec:problem_setup} and \ref{sec:failer_cases}, we discuss explanations for failure cases.
Here, we provide additional results.
Figure~\ref{fig:failure_more_class} and Fig.~\ref{fig:failure_more_box} show the explanations generated by each method for misclassification and mislocalization, respectively.
These results demonstrate the effectiveness of the proposed method in analyzing failure cases.
For example, 
as shown in the first row of Fig.~\ref{fig:failure_more_class},
where the object detector misclassifies the horse as a bird, 
the proposed method identifies not only the animal's body but also the tree outside the bounding box, indicating that these features (the animal's body and the tree) contribute to the misdetection of the bird.
In contrast, other methods do not highlight the tree outside the bounding box and fail to capture the effect of these features.
The proposed method is also effective in analyzing mislocalization.
For example, 
as shown in the first row of Fig.~\ref{fig:failure_more_box}, 
where the object detector mislocalizes the person by including the chair, 
the proposed method identifies both the features of the person and the towel covering the chair, indicating that these features contributed to the mislocalization.
In contrast, other methods either highlight only one of these features or produce noisy explanations.

\begin{table}[t]
  \centering
  \caption{Results of AUC for insertion (Ins), deletion (Del), overall (OA) metrics. The best values are indicated in bold.}
  \label{tab:del_ins_vps}
  \begin{tabular}{l|cccccc}
    \toprule
    \multicolumn{1}{c}{Detector} & \multicolumn{3}{c}{DETR} & \multicolumn{3}{c}{Faster R-CNN} \\
    \cmidrule(lr){2-4}\cmidrule(lr){5-7}
    \multicolumn{1}{c}{Metric} & Ins$\uparrow$ & Del$\downarrow$ & OA$\uparrow$ & Ins$\uparrow$ & Del$\downarrow$ & OA$\uparrow$ \\
    \midrule
    VPS      & \textbf{.908} & .088 & .820 & .901 & .120 & .781 \\
    VX-CODE  & .900 & \textbf{.058} & \textbf{.842} & \textbf{.912} & \textbf{.070} & \textbf{.842} \\
    \bottomrule
  \end{tabular}
\end{table}

\subsection{Comparison with VPS}
\label{sec:eval_vps}
We compare the proposed method with VPS~\citep{vps}, 
which was recently introduced for object-level foundation models, 
using DETR and Faster R-CNN.
Table~\ref{tab:del_ins_vps} presents the insertion (Ins), deletion (Del), and overall (OA) results obtained by applying both methods to these models.
Each value in \cref{tab:del_ins_vps} represents the average AUC over 500 instances with predicted class scores above 0.7 on MS-COCO.
We use VX-CODE with $r=1$.
The proposed method outperforms VPS across almost all metrics, demonstrating its effectiveness.

\begin{figure*}[ht]
    \centering
    \includegraphics[width=\linewidth]{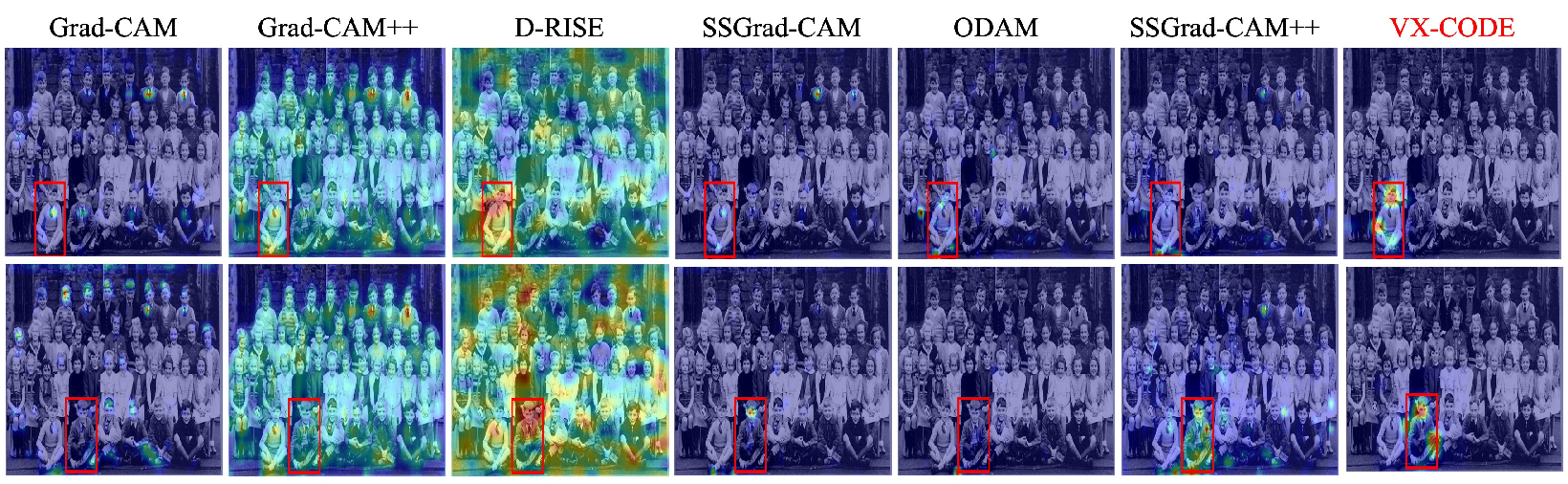}
    \caption{Heat maps for an image containing multiple instances of the same category. VX-CODE heat maps are generated via patch insertion.}
    \label{fig:compare_multi}
\end{figure*}

\subsection{Heat maps comparison with multiple instances}
\label{sec:heatmap_multi}
A key difference between detector explanations and classification explanations is that the former should be instance-specific.
To better demonstrate this property, 
we additionally provide qualitative results on images containing multiple objects from the same category.
Specifically, we consider cases where the detector outputs several bounding boxes of the same class and visualize explanations for a selected detection.

\par
Figure~\ref{fig:compare_multi} shows the results.
VX-CODE yields instance-focused evidence by capturing multiple complementary cues (e.g., the head and arms) that support the target detection.
In contrast, several baselines tend to produce noisy or ambiguous explanations that are harder to interpret in such multi-instance scenes.

\begin{figure*}[ht]
    \centering
    \includegraphics[width=\linewidth]{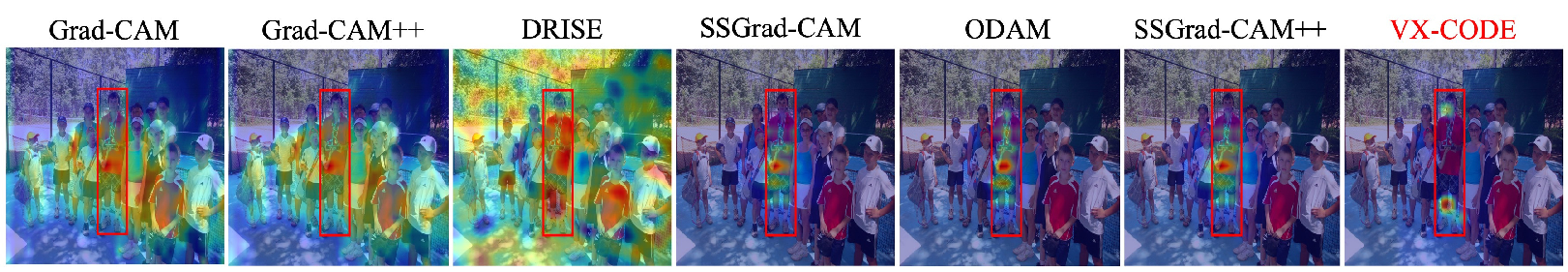}
    \caption{Heat map comparisons for RetinaNet. VX-CODE heat maps are generated via patch insertion.}
    \label{fig:heatmap_retina}
\end{figure*}

\begin{table}[t]
\centering
\caption{Quantitative results on RetinaNet (100 detected instances from COCO). 
The best values are indicated in bold.}
\label{tab:ins_del_retina}
\begin{tabular}{lccc}
\toprule
\textbf{Method} & \textbf{Ins} $\uparrow$ & \textbf{Del} $\downarrow$ & \textbf{OA} $\uparrow$ \\
\midrule
Grad-CAM        & 0.890 & 0.420 & 0.470 \\
Grad-CAM++      & 0.899 & 0.407 & 0.492 \\
DRISE           & 0.825 & 0.433 & 0.392 \\
SSGrad-CAM      & 0.924 & 0.369 & 0.555 \\
ODAM            & 0.929 & 0.354 & 0.575 \\
SSGrad-CAM++    & 0.932 & 0.348 & 0.584 \\ \midrule
VX-CODE         & \textbf{0.933} & \textbf{0.331} & \textbf{0.602} \\
\bottomrule
\end{tabular}
\end{table}

\subsection{Results on a one-stage detector}
\label{sec:one_stage}
In the main paper, we focus on two-stage, transformer-based detectors.
To further verify that VX-CODE generalizes beyond this setting,
we additionally evaluate it on the one-stage detector RetinaNet~\cite{focal_loss}.

\par
Figure~\ref{fig:heatmap_retina} compares the heat map explanations produced by each method.
VX-CODE also yields meaningful evidence regions on RetinaNet.
Table~\ref{tab:ins_del_retina} reports quantitative results computed on 100 detected instances from COCO.
Overall, VX-CODE outperforms the baselines, demonstrating its effectiveness on a one-stage detector.

\begin{table}[t]
\centering
\caption{Average pairwise interaction score $I$ on COCO with DETR (300 instances).
Higher values indicate stronger collective effects between selected patches.}
\label{tab:pairwise_interaction}
\begin{tabular}{lc}
\toprule
\textbf{Method} & \textbf{Interaction score} $\uparrow$ \\
\midrule
Grad-CAM        & 0.007 \\
Grad-CAM++      & 0.012 \\
DRISE           & 0.031 \\
SSGrad-CAM      & 0.017 \\
ODAM            & 0.020 \\
SSGrad-CAM++    & 0.025 \\ \midrule
VX-CODE         & \textbf{0.051} \\
\bottomrule
\end{tabular}
\end{table}

\subsection{Additional analysis of collective effects}
\label{sec:collective_effects}

We further analyze whether the selected patches exhibit collective effects.
For two selected patches $i$ and $j$, we define the pairwise interaction score as
\begin{equation}
I = f(\{i,j\}) - f(\{i\}) - f(\{j\}) + f(\emptyset),
\end{equation}
where $f$ denotes the reward function.
This corresponds to the interaction term in Def.~\ref{def:interaction} when $N=\{i,j\}$,
and measures the synergy obtained by inserting patches $i$ and $j$ jointly
into an empty image.

\par
For each method, we aggregate the explanation heat map into patch-level scores,
select the top-5 patches, and compute $I$ for all patch pairs.
We evaluate this on COCO using DETR over 300 instances.
Table~\ref{tab:pairwise_interaction} reports the average interaction score.

\par
VX-CODE achieves the highest interaction score, 
indicating that it tends to select patches with stronger collective effects than the competing methods.
This result complements the Del/Ins evaluation in the main paper and supports
the effectiveness of VX-CODE in identifying jointly important evidence.

\end{document}